\documentclass[10pt,twocolumn,letterpaper]{article}

\usepackage{3dv}
\usepackage{times}
\usepackage{epsfig}
\usepackage{graphicx}
\usepackage{amsmath}
\usepackage{amssymb}

\DeclareMathOperator*{\argmin}{argmin}
\DeclareMathOperator*{\soft}{soft}
\usepackage{xcolor}
\usepackage{stmaryrd}
\DeclareMathOperator{\median}{median}
\DeclareMathOperator{\diag}{diag}
\usepackage[pagebackref=true,breaklinks=true,letterpaper=true,colorlinks,bookmarks=false]{hyperref}

\threedvfinalcopy 

\def\threedvPaperID{291} 

\usepackage[toc,page,titletoc]{appendix}
\usepackage[capitalise,nameinlink]{cleveref}

\crefname{supp}{Supplement}{Supplements}

\begin{document}

\title{DeepBBS: Deep Best Buddies for Point Cloud Registration}

\author{Itan Hezroni~~~Amnon Drory~~~Raja Giryes~~~Shai Avidan\\
Tel Aviv University, Israel\\
{\tt\small \{itanhezroni, amnondrory\}@mail.tau.ac.il, \{raja, avidan\}@tauex.tau.ac.il}
}






\maketitle

\begin{abstract}
Recently, several deep learning approaches have been proposed for point cloud registration. These methods train a network to generate a representation that helps finding matching points in two 3D point clouds. Finding good matches allows them to calculate the transformation between the point clouds accurately. 
%
%
%
Two challenges of these techniques are dealing with occlusions and generalizing to objects of classes unseen during training. This work proposes DeepBBS, a novel method for learning a representation that takes into account the best buddy distance between points during training. Best Buddies (i.e., mutual nearest neighbors) are pairs of points nearest to each other. The Best Buddies criterion is a strong indication for correct matches that, in turn, leads to accurate registration.
Our experiments show improved performance compared to previous methods. In particular, our learned representation leads to an accurate registration for partial shapes and in unseen categories. Our code is publicly available\footnote{\url{https://github.com/itanhe/DeepBBS}}. 
\end{abstract}

\section{Introduction}

Rigid registration of point clouds is an important task in 3D shape processing. Given two scans of the same (rigid) object, the goal is to find the transformation that aligns one scan to the other. This has many applications in autonomous driving, 3D reconstruction, medical imaging, etc. In real life, scans are usually noisy and partial, so finding the correct transformation is hard. We consider here the case of rigid transformations with 6 degrees of freedom.


\begin{figure}
  \includegraphics[width=\linewidth]{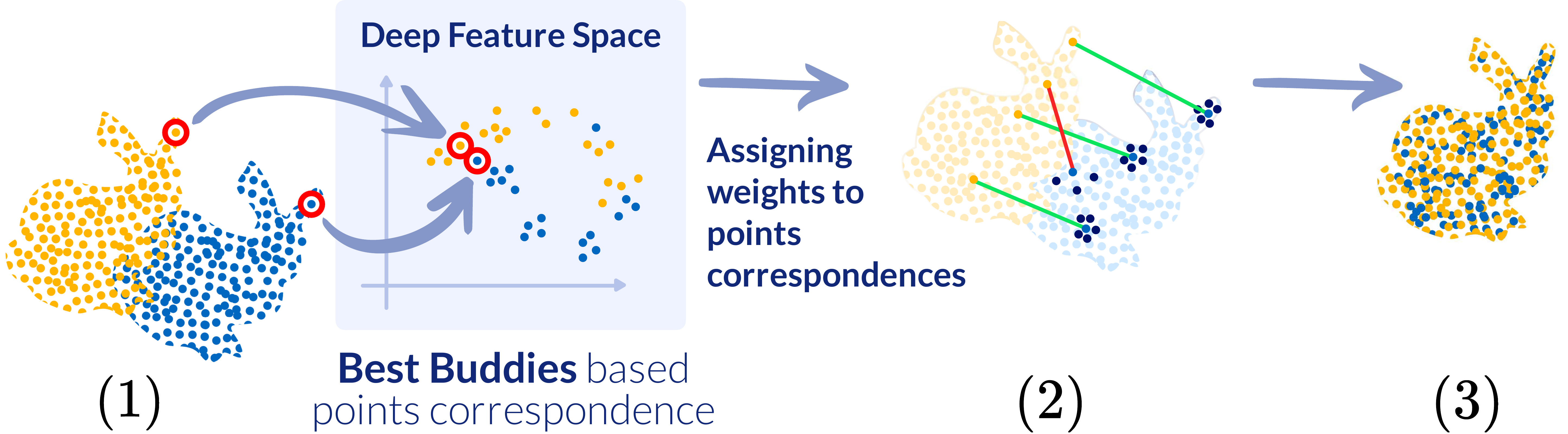}
\caption{\textbf{(1)} A corresponding pair of points that are far from each other in the 3D space are close to each other in a learned deep feature space. In this high dimensional space, we find for each point its best buddies in the other point cloud. \textbf{(2)} For every point in the orange point cloud, the corresponding points with the highest best buddy similarity (in the feature space) are marked in dark blue. A sum, weighted by a Best Buddy based measure, is computed to create a single corresponding point. The best buddy measure also weighs the correspondences (illustrated by green lines for good matches, and a red line for a bad match). \textbf{(3)} Then, the registration is performed by the weighted correspondences.}
  \label{fig:teaser}
\vspace{-0.1in}
\end{figure}

Perhaps the most common method is Iterative Closest Point (ICP) \cite{ICP}. It performs registration by first matching points between the point clouds and then finding the 6DoF transformation that minimizes the distance between the matched points. The process is iterated until convergence.



As matching is a challenging problem, several deep learning techniques have been recently proposed to tackle it. One of them, Deep Closest Point (DCP) \cite{DCP}, suggests a network that generates a representation for each point, and leads to an accurate matching between the point clouds. Yet, this approach struggles to provide a good representation when the point clouds are partial or from different classes than those used in training. 

To alleviate this problem, we propose the Deep Best Buddy Similarity learning (DeepBBS) strategy. It learns an embedding in which pairs of matching points are best buddies. A pair of points is called best buddies if each point is the nearest neighbor of the other (i.e., mutual nearest neighbors).
Since the best buddy measure is non-differentiable, we suggest a smooth differentiable version of it in the network, termed SoftBSS, which provides a larger weight for points that share a similar neighborhood. 

We trained a deep neural network that uses the SoftBBS component to weigh pairs of points. It provides a representation that can be used for matching complete and partial scans. The key stages of our method, denoted as DeepBBS, are illustrated in Figure~\ref{fig:teaser}. 
We test DeepBBS' performance on ModelNet40~\cite{ModelNet40} and on real scans~\cite{apollo_dataset, Turk:1994:ZPM}, showing its advantages both on complete and on partial shapes as well as on shape categories that were unseen during training.

\begin{figure*}
  \includegraphics[width=\textwidth]{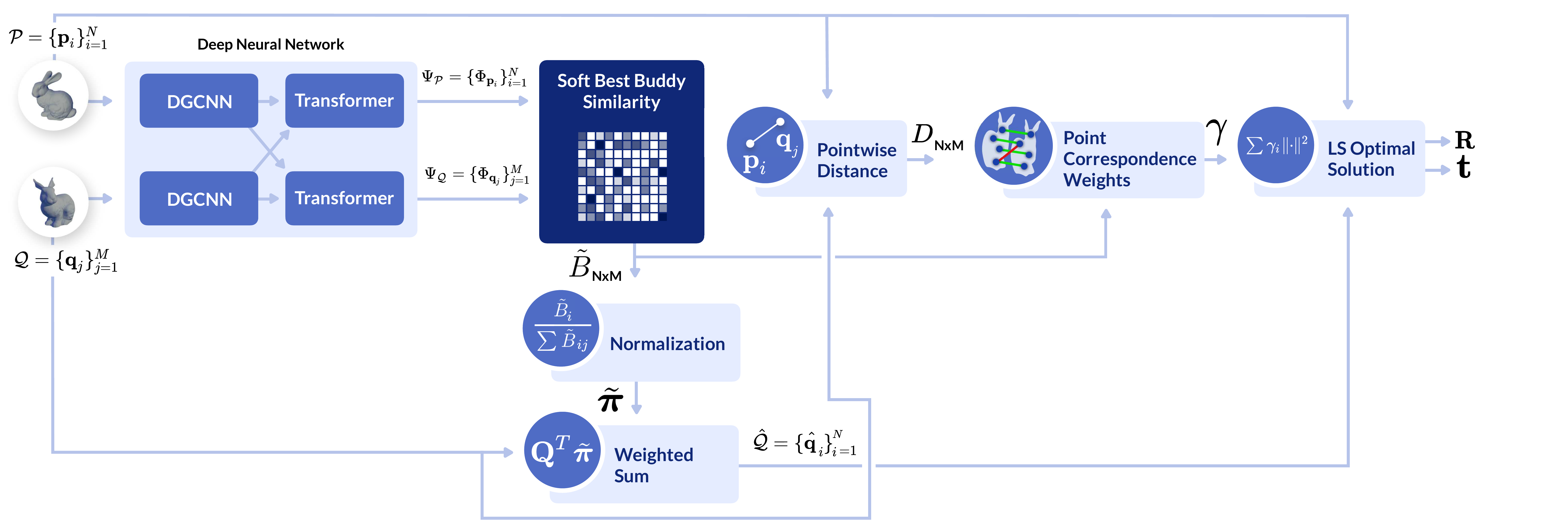}
\caption{\emph{DeepBBS algorithm overview:} the two point clouds are mapped to latent space in which we compute Best Buddy Pairs. That is, points that are the nearest neighbor of each other. These best buddy pairs are used to calculate the 3D rotation and translation between the two point clouds. The method is applied iteratively until convergence. }
\vspace{-0.15in}
  \label{fig:scheme}
\end{figure*}


\section{Related work}

We briefly survey works on point cloud registration of rigid objects, starting with classic techniques and then moving to discuss recent developments in deep learning.

One of the most prominent point clouds registration techniques is ICP \cite{ICP,Chen:1992}. ICP has been the workhorse of 3D point cloud registration, but it also suffers from several disadvantages. Particularly, it is sensitive to outliers and less accurate for partial matchings. Several outlier rejection methods and extensions have been proposed for ICP \cite{RUS:2001,Pomerleau:2015}. For example, Trimmed-ICP uses the least trimmed squares in the optimization to robustify it \cite{TrimmedICP}. In \cite{SparseICP}, a sparsity term is added to robustify the loss function to outliers and occlusions. 
Another strategy replaces the regular ICP minimization with the Levenberg-Marquardt method \cite{Fitzgibbon01c}. 

EM-ICP \cite{granger2002multi} applies ICP with multiple matching points assigned to each point instead of a single match. The associated points are used with Gaussian weights in the optimization for finding the transformation variables. KCReg \cite{tsin2004correlation} is an information theory based approach that measures the affinity between every pair of points using a kernel correlation. Their Renyi’s Quadratic Entropy based optimization objective measures the compactness of each point set for performing the registration. PM-SDP~\cite{maron2016point} uses convex relaxation with Procrustes Matching. MINA~\cite{bernard2020mina} matches points to a convex polyhedra for non-rigid shape matching.

Other works focus on the metric. Vanilla ICP uses a simple Euclidean distance to match points, but can be improved using a point-to-plane distance \cite{Chen:1992}. This has been extended to a full plane-to-plane distance in \cite{SegalHT09} and to a symmetric plane-to-plane distance \cite{Rusinkiewicz:2019:ASO}. The disadvantage of these extensions is that they require having the surface normals, which are not always provided and can be difficult to calculate, especially when the scanned point cloud is noisy.

Another line of works relies on using Gaussian distributions. One strategy represents each point cloud as a Gaussian mixture and finds the registration by minimizing a statistical discrepancy measure between the two mixtures \cite{Jian:2011:RPS}. This strategy was accelerated using a hierarchical Gaussian mixture representation (HGMR) \cite{Eckart_2018_ECCV}. Another acceleration was proposed in FilterReg \cite{Yang20TEASER}, which formulated the registration problem as a maximum likelihood optimization problem and then solved it iteratively using EM~\cite{FilterReg}. 

The Best Buddies Similarity (BBS) measure was introduced for robust template matching in images \cite{BBS}. It was also used in 2D for cross-domain correspondence \cite{aberman2018neural}. It was later used to register 3D point clouds~\cite{drory2020best}. The key idea was to define the (negative) BBS as a loss function and minimize it. We, on the other hand, rely on a neural network to learn a good representation for BBS. The approximation of BBS, which involves calculating nearest neighbors, relies on the soft approximation proposed in \cite{plotz2018neural} for image restoration. 




Various deep learning approaches have been proposed recently for registration. 
PointNetLK \cite{PointNetLK}, which relies on the PointNet architecture \cite{qi2017pointnet}, uses the Lucas-Kanade registration \cite{Lucas:1981} that is applied on a latent space calculated by PointNet for each point cloud. PCRNet~\cite{Sarode2019PCRNetPC} learns deep features of template and source point clouds to find the transformation that aligns them accurately. 
RPM-Net \cite{yew2020-RPMNet} and PointDSC \cite{Bai_2021_CVPR} further improve performance using deep features. Yet, they require having normal information.


PointGMM \cite{Hertz20PointGMM} and DeepGMR \cite{yuan2020deepgmr} learn to perform registration by representing the shapes via a hierarchical Gaussian mixture.
Deep Virtual Corresponding Points (DeepVCP) \cite{DeepICP} learns to select key-points that are matched probabilistically and then uses them calculate the transformation. 
Deep global registration (DGR) \cite{Choy_2020_CVPR} provides a confidence measure to matched points using a 6-dimensional convolutional network, then it uses the Procrustes algorithm to calculate the pose estimation, and finally it refines the final estimation by a robust gradient-based SE(3) optimizer.

The work that is most related to ours is Deep Closest Point \cite{DCP}. It proposes a neural network that learns a new representation for each point to improve the matching. It starts by finding an embedding for each point, then it uses a transformer-based approach to approximate the point matching, which is followed by a differentiable singular value decomposition (SVD) to approximate the 6DoF transformation parameters. PRNet \cite{PRNet} extended DCP to support partial scans by adding a key-point detection that finds shared points between two partial scans.

\section{Method}
\label{method}
We turn to describe our method. We start with a brief description of the registration setup, then present the best buddy similarity and propose using it as part of a registration network. This leads to a novel deep best buddy similarity learning (DeepBBS) approach, summarized in Figure \ref{fig:scheme}. It calculates the 6DoF parameters for the registration of two given input shapes. This method is robust to occlusions (getting partial shapes) and generalizes well.

\subsection{Setup}
\label{setup}
The goal of point cloud registration is to find a rigid transformation, i.e., a rotation and a translation which align a point cloud $\mathcal{P}=\{\mathbf{p}_i\}_{i=1}^{N}$ to a point cloud $\mathcal{Q}=\{\mathbf{q}_j\}_{j=1}^{M}$, where $\mathbf{p}_i, \mathbf{q}_j$ are points in ${\mathbb R}^3$. 
Following DCP~\cite{DCP}, we use the next minimization formula for the registration problem:
\begin{equation}
\label{eq:LSproblem}
    \mathbf{R},\mathbf{t} =  \argmin_{\substack{\mathbf{R} \in SO(3) \\ \mathbf{t} \in \mathbb{R}^3}} \left(\frac{1}{N} \sum_{i=1}^{N} \|\mathbf{R}\mathbf{p}_i + \mathbf{t} - \mathbf{q}_{\pi(\mathbf{p}_i)}\|^2 \right),
\end{equation}
where $\mathbf{R} \in SO(3)$ is a $3 \times 3$ rotation matrix, $\mathbf{t} \in {\mathbb R}^3$ is a $3 \times 1$ translation vector, and $\pi \colon {\mathbb R}^3 \to {\mathbb N}$ maps points in $\mathcal{P}$ to corresponding point indices in $\mathcal{Q}$. If the mapping $\pi$ is known, then this can be solved in closed-form:
\begin{equation}
    \mathbf{R} = \mathbf{V}\diag(1, 1, |\mathbf{VU}^T|) \mathbf{U}^T~~~~\mbox{and}~~~~~\mathbf{t} = -\mathbf{R}\mathbf{{\bar p}} + \mathbf{{\bar q}},
\end{equation}
where $\mathbf{U}$ and $\mathbf{V}$ are the left and right singular vectors of $\mathbf{H}=\mathbf{USV}^T$, with
\begin{equation}
\label{eq:cov}
    \mathbf{H} = \sum_{i=1}^{N} (\mathbf{p}_i-\mathbf{{\bar p}})(\mathbf{q}_{\pi(\mathbf{p}_i)}-\mathbf{{\bar q}}),
\end{equation}
and the centroids of $\mathcal{P}$ and $\mathcal{Q}$ are:
\begin{equation}
\label{eq:centroids}
    \mathbf{{\bar p}} = \frac{1}{N}\sum_{i=1}^{N} \mathbf{p}_i~~~~\mbox{and}~~~~\mathbf{{\bar q}} = \frac{1}{N}\sum_{i=1}^{N} \mathbf{q}_{\pi(\mathbf{p}_i)}.
\end{equation}
The challenge is finding the matching function $\pi$. ICP uses alternating minimization that calculates $\mathbf{R},\mathbf{t}$ given current $\pi$, and then updates $\pi$ given $\mathbf{R},\mathbf{t}$ by matching to each point in $\mathcal{P}$ its closest neighbor from $\mathcal{Q}$. Yet, this can lead to a wrong matching which may lead to a wrong registration. 

In general, matching 3D points is challenging as they do not have distinctive features. To face this problem, DCP \cite{DCP} learns deep features for each point so matching 3D points amounts to matching their corresponding deep features. The point clouds $\mathcal{P}$ and $\mathcal{Q}$ are mapped to point sets $\Psi_{\mathcal{P}}$ and $\Psi_{\mathcal{Q}}$ in a feature space by a Siamese DGCNN~\cite{DGCNN} followed by a Transformer module~\cite{Transformers}.

The proposed solution works well when there is an exact match between points in $\mathcal{P}$ and $\mathcal{Q}$, but performance degrades when this assumption breaks. In reality, this assumption is violated because more often than not, the same 3D surface is sampled more than once, and there is no reason to believe the exact same points will be sampled. Also, different 3D scans rarely cover the exact same region, so there are missing points due to occlusions or partial coverage.

\subsection{Best Buddy Similarity}

We extend DCP's~\cite{DCP} points matching by requiring that matching points are best buddies in the deep feature space. A pair of points in the embedding space $\mathbf{\Phi}_{\mathbf{p}_i} \in \Psi_{\mathcal{P}}$ and $\mathbf{\Phi}_{\mathbf{q}_j} \in \Psi_{\mathcal{Q}}$ are best buddies if $\mathbf{\Phi}_{\mathbf{p}_i}$ is closest to $\mathbf{\Phi}_{\mathbf{q}_j}$ and vice-versa. 
Formally, Best Buddy Similarity (BBS) is
\begin{equation}
\label{eq:bbp}
    B_{ij} = \llbracket i = \arg \min_{i'} D^{*}_{{i'}j} \rrbracket \cdot \llbracket j = \arg \min_{j'} D^{*}_{i{j'}} \rrbracket,
\end{equation}
where $D^{*}_{ij}=\left\|\mathbf{\Phi}_{\mathbf{p}_i}-\mathbf{\Phi}_{\mathbf{q}_j}  \right\|_2$ is the distance between $\mathbf{\Phi}_{\mathbf{p}_i}$ and $\mathbf{\Phi}_{\mathbf{q}_j}$, and $\llbracket \cdot \rrbracket$ is the indicator function that equals $1$ if the term in the brackets is true and zero otherwise. If $B_{ij}=1$ then $\mathbf{\Phi}_{\mathbf{p}_i}$ and $\mathbf{\Phi}_{\mathbf{q}_j}$ are called best buddies.

BBS is a non-differentiable metric because it uses the $\argmin$ operator. Thus, we use a differentiable approximation that we term SoftBBS, where we apply a $\soft \argmin$ operator. Specifically, $\tilde B$ approximates $B$ using:
\begin{equation}
\label{eq:softBBpairs}
    {{\tilde B}_{ij} = {\frac{e^{\frac{-D^{*}_{ij}}{\alpha}}}{\sum_{i'}e^{\frac{-D^{*}_{i'j}}{\alpha}}}}
    \cdot  {\frac{e^{\frac{-D^{*}_{ij}}{\alpha}}}{\sum_{j'}e^{\frac{-D^{*}_{ij'}}{\alpha}}}} }, \,\,\,\,\,\,\,\,\,
\end{equation}
where $\alpha$ is a temperature parameter. The matrix $\tilde B$ is the element-wise multiplication of row-wise and column-wise $\soft \argmin$ of the distance matrix $D^{*}$. Note the correspondence to the brackets in $B_{ij}$ definition. 

While $B_{ij}$ is non-zero only if $\mathbf{\Phi}_{\mathbf{p}_i}$ and $\mathbf{\Phi}_{\mathbf{q}_j}$ are mutual nearest neighbors, ${\tilde B}_{ij}$ can be non-zero when, for example, $\mathbf{\Phi}_{\mathbf{p}_i}$ is $\mathbf{\Phi}_{\mathbf{q}_j}$'s 3rd nearest neighbor, while $\mathbf{\Phi}_{\mathbf{q}_j}$ is $\mathbf{\Phi}_{\mathbf{p}_i}$'s 4th nearest neighbor. The value of the temperature parameter, $\alpha$, controls this behavior. The smaller it is, the more strict ${\tilde B}_{ij}$ becomes, i.e., similar to $B_{ij}$. We want $\frac{D^{*}_{ij}}{\alpha}\approx 1$ for best buddies, and $\frac{D^{*}_{ij}}{\alpha} \gg 1$ for points that are not best buddies. Since the best buddies are not known, we use the ``typical minimal distance between neighbours'' calculated as  $\xi \cdot \median_{j}\left(\min_{i}\left\|\mathbf{\Phi}_{\mathbf{p}_j}-\mathbf{\Phi}_{\mathbf{p}_i}\right\|\right)$, where $\xi$ is a constant factor. We found empirically that $\xi = 0.4$ leads to the best results, in which most points have only a few best buddies.

\subsection{Mapping Function and Weightening}
\label{mapping}
Given the BBS measure, we define the soft mapping
\begin{equation}
    \boldsymbol{\tilde\pi}(\mathbf{p}_i) = \frac{\tilde B_{i}}{\sum_{j=1}^{M}\tilde B_{ij}},
\end{equation}
where $\boldsymbol{\tilde\pi} \colon \mathbb{R}^{3} \to \mathbb{R}^{M}$ and $\tilde B_{i}$ is the $i$th row of $\tilde B$. $\boldsymbol{\tilde\pi}$ acts as a normalized pointer from $\mathbf{p}_i$ to its best buddies in $\mathcal{Q}$.

Given the mapping $\boldsymbol{\tilde\pi}$, we construct a point cloud $\mathcal{\hat Q}=\{\mathbf{\hat{q}}_i\}_{i=1}^{N}$ that matches $\mathcal{P}$ as follows:
\begin{equation}
\label{eq:calc_q_hat}
    \mathbf{{\hat q}}_i = \mathbf{Q}^T\boldsymbol{\tilde\pi}(\mathbf{p}_i) \in {\mathbb R}^3,
\end{equation}
where the rows of $\mathbf{Q} \in \mathbb{R}^{M \times 3}$ contain $\{\mathbf{q}_j\}_{j=1}^{M}$. This operation, illustrated in the supplementary material (\ref{supp:more_figures}), generates a mapping $\mathbf{p}_i \rightarrow \mathbf{{\hat q}}_i$. $\mathbf{{\hat q}}_i$ is generated by a weighted sum of points in $\mathcal{Q}$, weighted by $\boldsymbol{\tilde\pi}(\mathbf{p}_i)$, i.e., points that are the best buddies of $\mathbf{p}_i$ in $\mathcal{Q}$. The result of the weighted sum should fit $\mathbf{p}_i$. Ideally, $\mathbf{{\hat q}}_i$ and $\mathbf{p}_i$ differ only by the transformation between $\mathcal{P}$ and $\mathcal{Q}$. A visual example of such mapping appears in the supplementary material (\ref{supp:more_figures}). 

In order to reject unsuccessful matches or missing matches caused by occlusions, we assess the correspondence of the pair $(\mathbf{p}_i, \mathbf{\hat q}_i)$ with a weight $\gamma_i$ :
\begin{equation}
\label{eq:gamma_calc}
    \gamma_i = \sum_{j=1}^{M} {{\tilde B}_{ij}}e^{-D_{ij}/T},
\end{equation}
where ${\tilde B}_{ij}$ is the SoftBBS (see Equation~\eqref{eq:softBBpairs}),  $D_{ij}$ is the Euclidean distance between $\mathbf{p}_i$ and $\mathbf{q}_j$, and $T$ is a temperature parameter that is learned during training. The weight $\gamma_i$ is composed of the SoftBBS term that provides a robust matching mechanism, which captures similarities in the feature space, and from a spatial term that measures similarities in the input 3D space. The spatial term might unintentionally decrease $\gamma_i$ at the beginning of the registration process when the point clouds are far from each other. Thus, we decrease T at inference. Further discussion is in Section \ref{inference}. $\gamma_i$ sums over all of the points in $\mathcal{Q}$. Thus a pair $(\mathbf{p}_i, \mathbf{\hat q}_i)$ will get a high value of $\gamma_i$ if $\mathbf{\hat q}_i$ was constructed from points that are "good" best buddies and are close to $\mathbf{p}_i$. In the case that $\mathbf{p}_i$ has no matching points in $\mathcal{Q}$ due to occlusion, $\tilde{B}_{ij}$ and $e^{-D_{ij}/T}$ should be small $\forall j$, hence the correspondence parameter $\gamma_i$ should be small as well.

\subsection{The Weighted Problem}
Given the weights, $\gamma_i$, we propose a weighted version of the minimization problem in Equation~\eqref{eq:LSproblem}:
\begin{equation}
    \mathbf{R},\mathbf{t} =  \argmin_{\substack{\mathbf{R} \in SO(3) \\ \mathbf{t} \in \mathbb{R}^3}} \left(\frac{1}{N} \sum_{i=1}^{N} \gamma_i ||\mathbf{R}\mathbf{p}_i + \mathbf{t} - \mathbf{\hat q}_i||^2 \right).
\end{equation}
Its solution is very similar to Equation~\eqref{eq:LSproblem} \cite{LSSVF}, where here  
\begin{equation}
\label{eq:w_cov}
    \mathbf{H} = \sum_{i=1}^{N} \gamma_i (\mathbf{p}_i-\mathbf{{\bar p}})(\mathbf{{\hat q}}_i-\mathbf{{\bar {\hat q}}}),
\end{equation}
and
\begin{equation}
\label{eq:w_centroids}
    \mathbf{{\bar p}} = \frac{\sum_{i=1}^{N} \gamma_i \mathbf{p}_i}{\sum_{i=1}^{N} \gamma_i}~~~~\mbox{and}~~~~\mathbf{\bar {{\hat q}}} = \frac{\sum_{i=1}^{N} \gamma_i \mathbf{{\hat q}}_i}{\sum_{i=1}^{N} \gamma_i}.
\end{equation}


\subsection{Loss Function}
The loss that is used for training the network weights is
\begin{eqnarray}
\label{eq:loss}
    \mathcal{L}=&&\|\mathbf{R}_{GT}^T\mathbf{R}-\mathbf{I}\|^2 + 
    \|\mathbf{t}_{GT}-\mathbf{t}\|^2 \\ \nonumber &&
    + \beta^n\frac{1}{N}\sum_{i=1}^{N}\gamma_{GT,i}\left\|\mathbf{\hat q}_i-(\mathbf{R}_{GT}\mathbf{p}_i+\mathbf{t}_{GT})\right\|^2.
\end{eqnarray}
$\mathbf{R}_{GT}$ and $\mathbf{t}_{GT}$ are the ground truth rotation and translation (respectively), and $\mathbf{R}$ and $\mathbf{t}$ are the predicted ones. $\beta$ is a decay constant, and $n$ is the epoch number. 
$\gamma_{GT,i}$ equals $1$ if $\mathbf{p}_i$ has a matching point in $\mathcal{Q}$'s coordinate system within a $\theta$ threshold, i.e.: $\gamma_{GT,i}=\llbracket\min_j\left\|\mathbf{q}_j-(\mathbf{R}_{GT}\mathbf{p}_i+\mathbf{t}_{GT})\right\|^2<\theta\rrbracket$. We used $\theta=0.02$ and $\beta=0.95$.

The first two terms in $\mathcal{L}$ penalize an inaccurate transformation estimation. The third term requires $\mathbf{\hat q}_i$ to be close to $\mathbf{p}_i$ when $\mathbf{p}_i$ is transformed with the ground truth transformation. As mentioned in Section~\ref{mapping}, $\mathbf{p}_i$ should be equal to $\mathbf{\hat q}_i$ in $\mathcal{Q}$'s coordinate system. The third term makes the network learn weights that generate a more accurate point mapping, $\boldsymbol{\tilde\pi}(\mathbf{p}_i)$. It does not depend on the predicted $\mathbf{R}$ and $\mathbf{t}$. Thus, it serves as a skip connection in backpropagation as gradients pass through the LS solution. As a result, the learning process converges faster. The term's weight decays, as we want accuracy in the predicted transformation to be more dominant as the learning process progresses.

\subsection{Inference}
\label{inference}
We consider two variants of our algorithm, termed DeepBBS and DeepBBS++.

{\bf DeepBBS.} At inference time, we estimate $\mathbf{R}$ and $\mathbf{t}$ with our algorithm. Then, we apply the estimated transformation on $\mathcal{P}$ for getting $\mathcal{P}'$. We iterate the algorithm on $\mathcal{Q}$ and $\mathcal{P}'$ until the rotation angle difference between consequent iterations is less than a threshold ($0.4$ in our tests). This process typically takes 2-3 iterations. The temperature parameter $T$, which balances the similarity in the feature space and the 3D input space, is reduced by a factor of $2$ every iteration. This gives more weight to similarities in the 3D space as the point clouds align with each other. Thus, it helps the method lock on to the correct matches and to converges to a better result. This is opposed to the beginning of the registration process when less weight is given to similarities in the 3D space when the point clouds are far from each other because $T$ is relatively large. 

{\bf DeepBBS++.} After getting $\mathbf{R}$ and $\mathbf{t}$ using DeepBBS, we fine-tune the result by iterating in the 3D space. We do so because we found that when the point clouds are almost aligned, and only fine-tuning is required, finding the best buddies based on similarity in the 3D space is more accurate than in the feature space. To do so, we skip the deep neural network part, i.e., $\tilde{B}$ is calculated with the 3D points ($\mathcal{P}$ and $\mathcal{Q}$ as input points instead of the points in the feature space ($\Psi_{\mathcal{P}}$ and $\Psi_{\mathcal{Q}}$)). Here we use $T=1$. In an ablation study (in the supplementary material \ref{supp:ablation_modelnet40}) it is shown that applying only the fine-tuning step yields inaccurate results.

\begin{table*}
\begin{center}
\begin{tabular}{|l|c|c|c|c|c|c|}
\hline
Method & MSE($\mathbf{R}$) & RMSE($\mathbf{R}$) & MAE($\mathbf{R}$) &
         MSE($\mathbf{t}$) & RMSE($\mathbf{t}$) & MAE($\mathbf{t}$) \\
\hline\hline
ICP~\cite{ICP}          & $1134.552$ & $33.683$ & $25.045$ & $0.0856$ & $0.293$ & $0.250$ \\
Go-ICP~\cite{SegalHT09} & $195.985$  & $13.999$ & $3.165$  & $0.0011$ & $0.033$ & $0.012$ \\ 
FGR~\cite{FGR}        & $126.288$  & $11.238$ & $2.832$  & $0.0009$ & $0.030$ & $0.008$ \\
SymmetricICP~\cite{Rusinkiewicz:2019:ASO}        & $412.572$  & $20.312$ & $6.242$  & $0.0081$ & $0.090$ & $0.037$ \\
PointNetLK~\cite{PointNetLK} & $280.044$  & $16.375$ & $7.550$  & $0.0020$ & $0.045$ & $0.025$ \\
DCP-v2~\cite{DCP}     & $45.005$   & $6.709$  & $4.448$  & $0.0007$ & $0.027$ & $0.020$ \\
PRNet~\cite{PRNet}    & $10.235$   & $3.199$  & $1.454$  & $0.0003$ & $0.016$ & $0.010$ \\
\hline
DeepBBS (ours) & $0.002$ & $0.041$& $0.021$  & $0.000001$ & $0.0007$ & $0.0004$ \\
DeepBBS++ (ours) & $\mathbf{0.0002}$ & $\mathbf{0.014}$& $\mathbf{0.006}$  & $\mathbf{0.0000001}$ & $\mathbf{0.0004}$ & $\mathbf{0.0001}$ \\
\hline
\end{tabular}
\end{center}
\caption{Results on unseen point clouds. Our method outperforms all others.}
\label{table:unseen_point_clouds}
\end{table*}

\begin{table*}
\begin{center}
\begin{tabular}{|l|c|c|c|c|c|c|}
\hline
Method & MSE($\mathbf{R}$) & RMSE($\mathbf{R}$) & MAE($\mathbf{R}$) &
         MSE($\mathbf{t}$) & RMSE($\mathbf{t}$) & MAE($\mathbf{t}$) \\
\hline\hline
ICP~\cite{ICP}          & $1217.618$ & $34.894$ & $25.455$ & $0.086$ & $0.293$ & $0.251$ \\
Go-ICP~\cite{SegalHT09} & $157.072$  & $12.533$ & $2.940$  & $0.0009$ & $0.031$ & $0.010$ \\ 
FGR~\cite{FGR}          & $98.635$  & $9.932$ & $1.952$  & $0.0014$ & $0.038$ & $0.007$ \\
SymmetricICP~\cite{Rusinkiewicz:2019:ASO}        & $363.195$  & $19.058$ & $5.847$  & $0.0081$ & $0.090$ & $0.038$ \\
PointNetLK~\cite{PointNetLK} & $526.401$  & $22.943$ & $9.655$  & $0.0037$ & $0.061$ & $0.033$ \\
DCP-v2~\cite{DCP}     & $95.431$   & $9.769$  & $6.954$  & $0.0010$ & $0.034$ & $0.025$ \\
PRNet~\cite{PRNet}    & $24.857$   & $4.986$  & $2.329$  & $0.0004$ & $0.021$ & $0.015$ \\
PRNet*~\cite{PRNet}   & $15.624$   & $3.953$  & $1.712$  & $0.0003$ & $0.017$ & $0.011$ \\
\hline
DeepBBS (ours) & $0.006$ & $0.075$& $0.040$  & $0.000001$ & $0.0011$ & $0.0006$ \\
DeepBBS++ (ours) & $\mathbf{0.0006}$ & $\mathbf{0.024}$& $\mathbf{0.008}$  & $\mathbf{0.0000002}$ & $\mathbf{0.0005}$ & $\mathbf{0.0002}$ \\
\hline
\end{tabular}
\end{center}
\caption{Results on unseen categories. Our method outperforms all others.}
\label{table:unseen_categories}
\end{table*}

\begin{table*}
\begin{center}
\begin{tabular}{|l|c|c|c|c|c|c|}
\hline
Method & MSE($\mathbf{R}$) & RMSE($\mathbf{R}$) & MAE($\mathbf{R}$) &
         MSE($\mathbf{t}$) & RMSE($\mathbf{t}$) & MAE($\mathbf{t}$) \\
\hline\hline
ICP~\cite{ICP}          & $1229.670$ & $35.067$ & $25.564$ & $0.0860$ & $0.294$ & $0.250$ \\
Go-ICP~\cite{SegalHT09} & $150.320$  & $12.261$ & $2.845$  & $0.0008$ & $0.028$ & $0.029$ \\ 
FGR~\cite{FGR}          & $764.671$  & $27.653$ & $13.794$  & $0.0048$ & $0.070$ & $0.039$ \\
SymmetricICP~\cite{Rusinkiewicz:2019:ASO}        & $428.068$  & $20.690$ & $6.773$  & $0.0085$ & $0.092$ & $0.044$ \\
PointNetLK~\cite{PointNetLK} & $397.575$  & $19.939$ & $9.076$  & $0.0032$ & $0.057$ & $0.032$ \\
DCP-v2~\cite{DCP}     & $47.378$   & $6.883$  & $4.534$  & $0.0008$ & $0.028$ & $0.021$ \\
PRNet~\cite{PRNet} & $18.691$ & $4.323$& $2.051$  & $\mathbf{0.0003}$ & $\mathbf{0.017}$ & $\mathbf{0.012}$ \\
\hline
DeepBBS (ours)     & $17.625$   & $4.198$  & $1.715$  & $0.0019$ & $0.0441$ & $0.023$ \\
DeepBBS++ (ours)     & $\mathbf{16.568}$   & $\mathbf{4.070}$  & $\mathbf{0.974}$  & $0.0022$ & $0.0471$ & $0.017$ \\
\hline
\end{tabular}
\end{center}
\caption{Results on point clouds corrupted with white Gaussian noise. Our method is best in rotation error and second in MAE($\mathbf{t}$).}
\label{table:noise}
\end{table*}

\begin{figure}[t]
  \includegraphics[width=\linewidth]{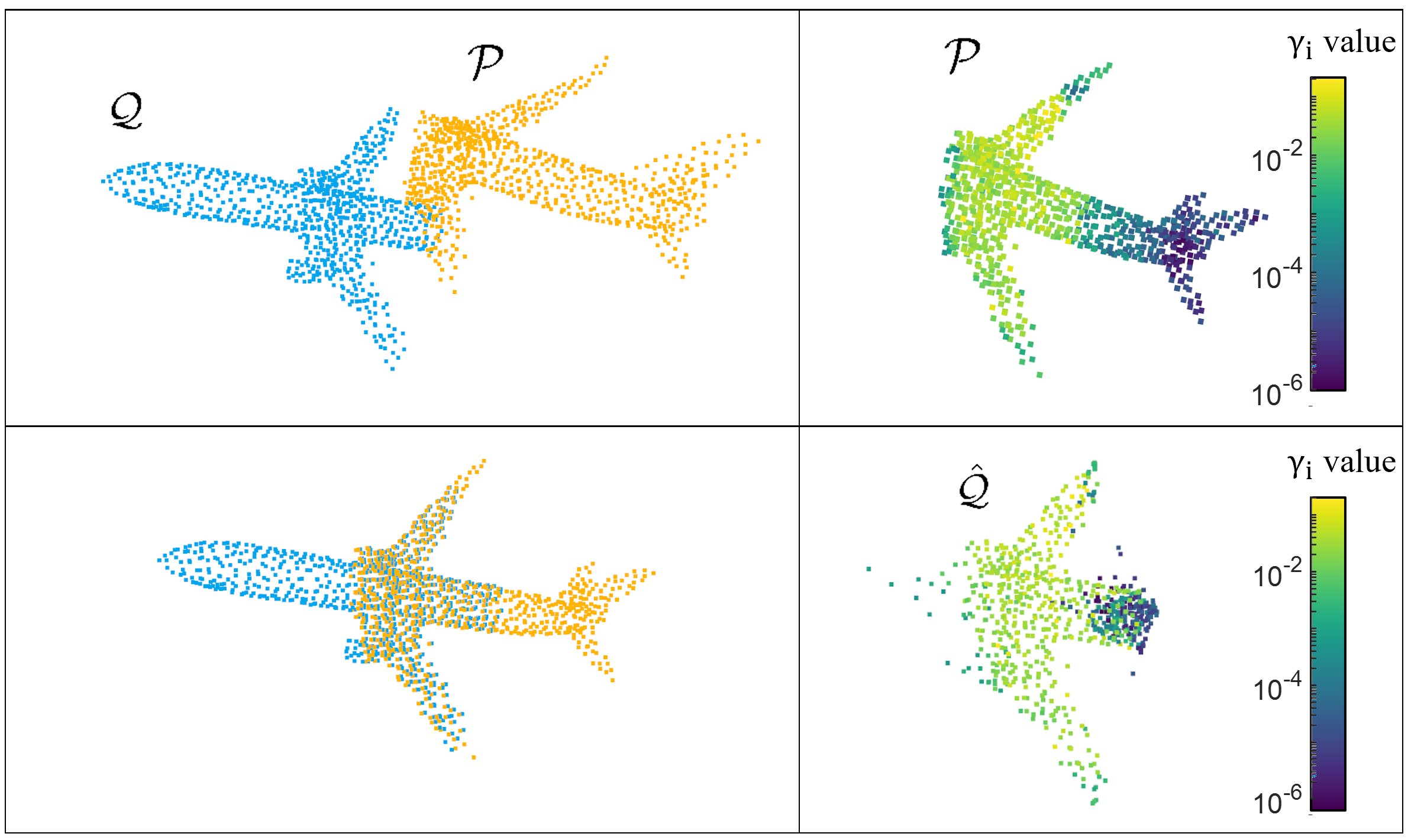}
\caption{\emph{Example of dealing with partial scanning.} \textbf{Top-Left:} $\mathcal{P}$ and $\mathcal{Q}$, in orange and blue, respectively. Only the middle part of the airplane is mutual between the scans. \textbf{Top-Right:} point cloud $\mathcal{P}$. $\mathbf{p}_i$ is colored with $\gamma_i$ (on a log scale). Points of the airplane tail, which have no matches in $\mathcal{Q}$, receive a low value of $\gamma$, which yields no influence on the result. \textbf{Bottom-Right:} The modified point cloud $\hat{\mathcal{Q}}$. $\hat{\mathbf{q}}_i$ is colored with $\gamma_i$ (on a log scale). Points that do not fit the airplane shape (outliers or ones with no corresponding points in the other shape) receive a low value of $\gamma$. \textbf{Bottom-Left:} Final registration result.}
  \label{fig:partial_gamma}
\end{figure}

\begin{figure}[t]
\includegraphics[width=\linewidth]{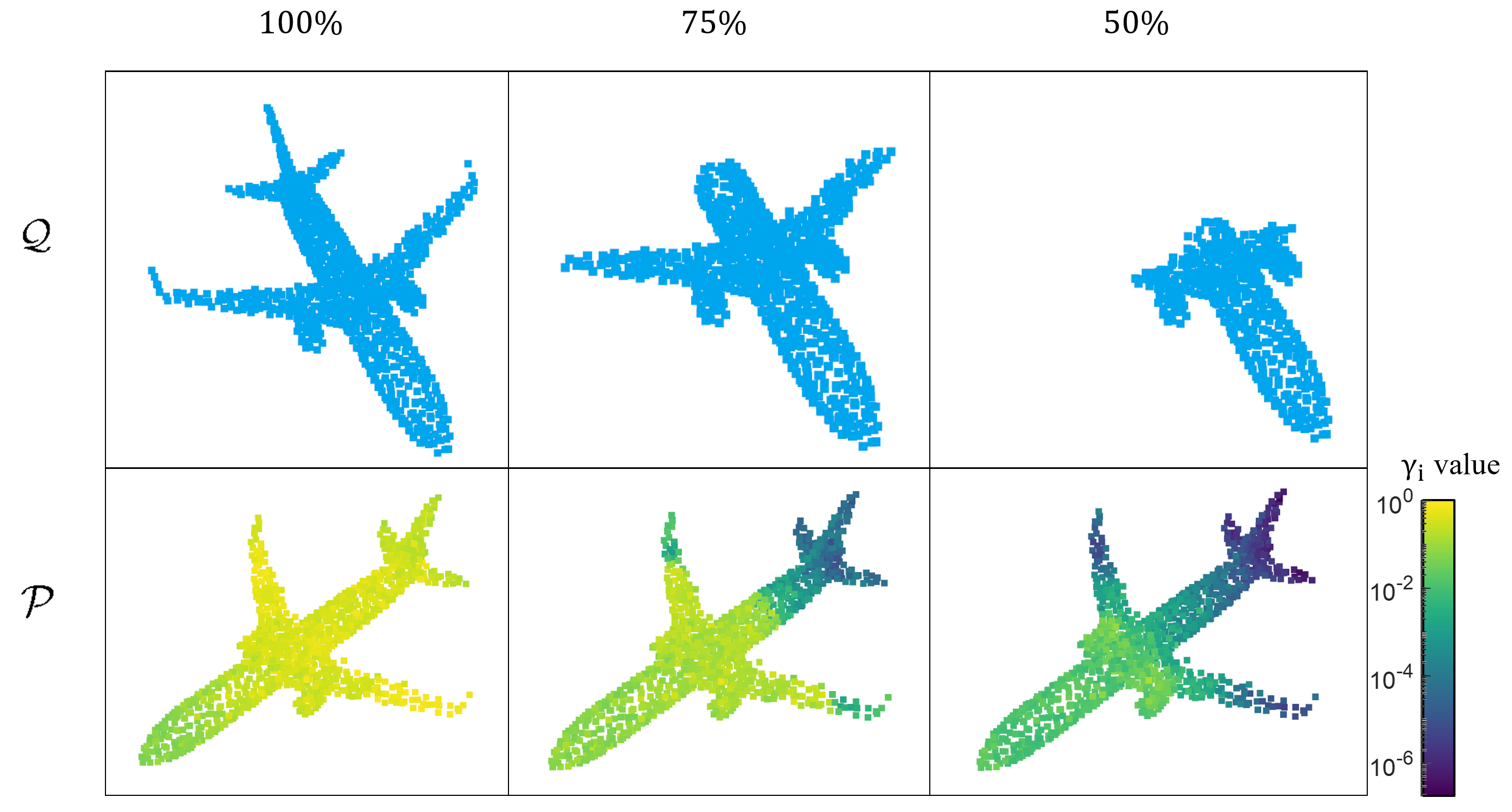}
\caption{\emph{Occlusions detection by $\gamma$.} In the first row, a point cloud $\mathcal{Q}$ with an increasing amount of occlusions. In the second row, a point cloud $\mathcal{P}$ colored with $\gamma_i$ values. In the case of no occlusions, all points in $\mathcal{P}$ have high $\gamma_i$ values. When there are occlusions, points with no corresponding point in $\mathcal{Q}$ have low $\gamma_i$ values. Hence, they will have a negligible  influence on the registration.}
\label{fig:different_subsampling}
\end{figure}

\section{Experiments}

We use several experiments to evaluate our method. First, we follow some of the experiments of \cite{PRNet} conducted on the ModelNet40~\cite{ModelNet40} dataset. We then perform a new test that was designed on this dataset. Finally, we evaluate our algorithm on two real datasets - the Stanford Bunny~\cite{Turk:1994:ZPM} and Apollo-SouthBay~\cite{apollo_dataset} datasets.

The network was trained using the Adam~\cite{kingma2014adam} optimizer for $250$ epochs. The initial learning rate was set to $0.001$ and was decreased by $10$ after $130$, $200$ and $225$ epochs. We used $4$ Nvidia TITAN X GPUs for training.


\subsection{ModelNet40 Dataset}
\label{modelnet40}
ModelNet40~\cite{ModelNet40} contains $40$ categories of 3D objects. It is divided into train and test sets. In this experiment, each shape is represented using $1024$ points sampled using the farthest-point sampling from the original CAD model.
Each point cloud is transformed using a 3D rigid transformation. We use a random rotation of up to 45$^{\circ}$ around each axis and a random translation in the range of $[-0.5, 0.5]$ in each axis. Since the transformation is artificial, the learning is unsupervised.
We present $4$ experiments. In the first $3$ experiments, our method is evaluated on two partial scans of an object. In the 4'th experiment the point clouds have a different sampling. To simulate a partial-to-partial registration between the shapes $\mathcal{P}$ and $\mathcal{Q}$, a random 3D point from each point cloud is selected, and $768$ nearest neighbor points are sampled. This creates two point clouds with a partial overlap. In these experiments, the overlapping points are {\em exact} matches in the two point clouds.

The authors of~\cite{PRNet} compare several methods and we report their findings here. The methods that are evaluated include ICP~\cite{ICP}, Go-ICP~\cite{SegalHT09}, FGR~\cite{FGR}, PointNetLK~\cite{PointNetLK}, DCP~\cite{DCP}, and PRNet~\cite{PRNet}. The first three methods are classical, while the last three use Deep Learning. Interestingly, FGR~\cite{FGR} uses a best-buddies test (termed "Reciprocity test") to initialize their point matches. The test is carried out on the Fast Point Feature Histogram (FPFH) feature~\cite{FPFH}, which is a hand-crafted feature, as opposed to the deep features used in our method. 
We also test Symmetric-ICP of Rusinkiewicz~\cite{Rusinkiewicz:2019:ASO}, which is not learning-based.

We evaluate our performance and compare it to other techniques by measuring MAE (Mean Absolute Error), MSE (Mean Square Error) and RMSE (Root MSE) of the predicted Euler angles and the predicted translation. Visual results appear in the supplementary material (\ref{supp:more_visual_exp}).

\noindent {\bf Unseen partial point clouds.}
The first experiment measures partial-to-partial registration of unseen point clouds during training. The dataset consists of 40 classes that are split into $80\%/20\%$ train/test split. %
Table~\ref{table:unseen_point_clouds} reports the results showing that our method outperforms all others.

Figure \ref{fig:partial_gamma} demonstrates the role of $\gamma$ in filtering points. We show its values both on the point cloud $\mathcal{P}$ and the modified point cloud $\hat{\mathcal{Q}}$.  The shapes' overlapping parts receive the largest values of $\gamma$, showing how the network focuses on them to calculate the transformations. This leads to an accurate registration of the partial shapes. Figure~\ref{fig:different_subsampling} shows that points with high $\gamma$ values in the case of no occlusions, get low values when their matching points are missing. $\gamma$ was calculated with the network's weights of this experiment.

\noindent {\bf Unseen categories of partial point clouds.}
Also following~\cite{PRNet}, we test the different algorithms' generalization performance on unseen categories: We train on 20 classes of ModelNet40~\cite{ModelNet40} and evaluate the remaining 20 classes.
Table~\ref{table:unseen_categories} reports the results. We see that here, as well, our methods outperform all other techniques.

\noindent {\bf Noisy partial scans.}
We also use point clouds corrupted with white Gaussian noise. Table~\ref{table:noise} shows that we outperform all methods, except for~\cite{PRNet}, in its translation error.

\begin{table*}
\begin{center}
\begin{tabular}{|l|c|c|c|c|c|c|}
\hline
Method & MSE($\mathbf{R}$) & RMSE($\mathbf{R}$) & MAE($\mathbf{R}$) &
         MSE($\mathbf{t}$) & RMSE($\mathbf{t}$) & MAE($\mathbf{t}$) \\
\hline\hline
SymmetricICP~\cite{Rusinkiewicz:2019:ASO}        & $247.682$  & $15.738$ & $4.576$  & $0.0017$ & $0.042$ & $0.014$ \\
BBR~\cite{drory2020best}    & $739.554$  & $27.195$ & $16.746$ & $0.004533$ & $0.213$ & $0.162$ \\
BD~\cite{drory2020best}     & $827.459$  & $28.765$ & $24.549$ & $0.08245$ & $0.287$ & $0.249$ \\
BDN~\cite{drory2020best}    & $672.882$  & $25.940$ & $22.407$ & $0.08250$ & $0.287$ & $0.249$ \\
PRNet~\cite{PRNet}    & $14.833$   & $3.851$  & $1.821$  & $0.00020$ & $0.014$ & $0.011$ \\
\hline
DeepBBS (ours) & $\mathbf{8.566}$ & $\mathbf{2.927}$& $\mathbf{1.089}$  & $\mathbf{0.00006}$ & $\mathbf{0.008}$ & $\mathbf{0.006}$ \\
DeepBBS++ (ours) & $11.914$ & $3.452$& $1.640$  & $0.00044$ & $0.021$ & $0.016$ \\
\hline
\end{tabular}
\end{center}
\caption{Results on point clouds of different samplings. Our method outperforms all others.}
\label{table:bbs_comparisons}
\end{table*}

\begin{figure}
  \includegraphics[width=\linewidth]{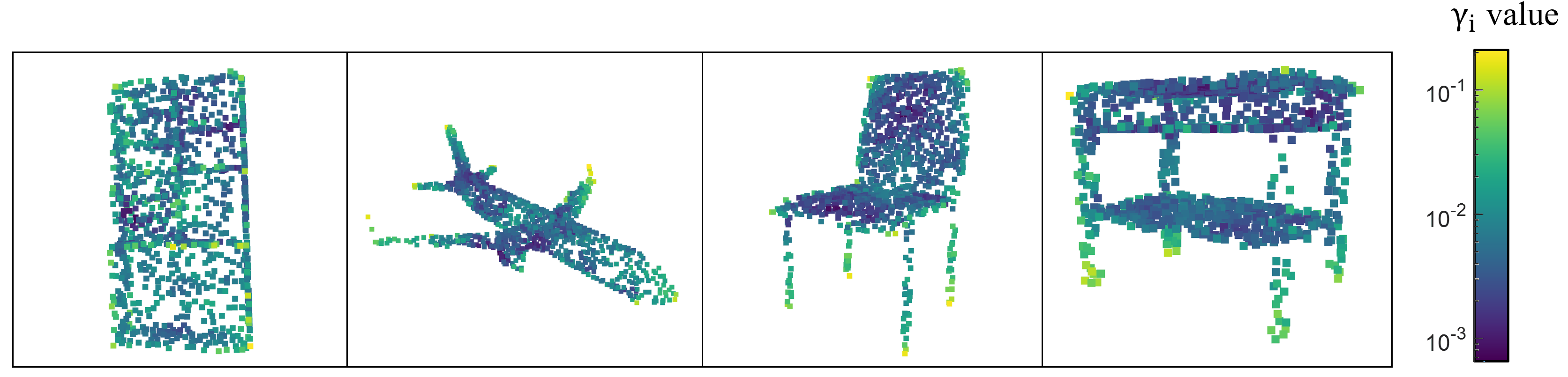}
\caption{Each panel shows a different point cloud $\mathcal{P}$ from ModelNet40, taken from the different sampling experiment. Each point $\mathbf{p}_i$ is colored with $\gamma_i$. Notice that corners and edges, which are distinct points and are helpful for registration, tend to get high $\gamma$ values, while points on planes tend to get low values of $\gamma$.}
  \label{fig:gamma}
\end{figure}

\noindent {\bf Different sampling.}
\label{different_sampling}
We test registration in the case of different point samplings of the point clouds, i.e., we randomly sampled $1024$ different points in $\mathcal{P}$ and $\mathcal{Q}$. In this test, we compare our method to PRNet~\cite{PRNet}, SymmetricICP~\cite{Rusinkiewicz:2019:ASO}, as well as three versions of~\cite{drory2020best}. The method of~\cite{drory2020best} also uses BBS but with two important distinctions. First, it optimizes the rigid transformation as part of the back-propagation, as opposed to our method that {\em estimates} the parameters. Thus, \cite{drory2020best} requires hundreds of iterations to converge while our method takes, on average, only 3. Second, \cite{drory2020best} uses best-buddy similarity in the input space, while we measure best-buddy similarity in the latent feature space. 

Table~\ref{table:bbs_comparisons} shows that {\bf DeepBBS} outperform other methods. Figure \ref{fig:gamma} shows examples of point clouds $\mathcal{P}$, taken from this experiment. The color of each point $\mathbf{p}_i$ is determined by the value of its corresponding weight $\gamma_i$. The key role of $\gamma$ in finding key points for registration is demonstrated by the observation that distinct points get high values of $\gamma$, hence they are dominant in the estimation of the transformation.

An ablation study evaluating the contributions of the different components of {\bf DeepBBS++} can be found in the supplementary material (\ref{supp:ablation_modelnet40}). In short, we found that all of the components of our method improve the results.

\subsection{Stanford Bunny Dataset}
\label{stanfordbunny}
We test our method on real scans of the Stanford Bunny dataset~\cite{Turk:1994:ZPM}. It consists of $10$ scans of a bunny model taken from different angles. We follow Rusinkiewicz's~\cite{Rusinkiewicz:2019:ASO} framework, containing $19$ pairs of scans having IOU overlap greater than $20\%$. $1000$ points from each scan are sampled randomly, and then the point clouds are aligned to each other. A rotation of $0^{\circ}-60^{\circ}$ around a random axis and a translation of $0\%-50\%$ of the model size in a random direction are performed. The test is repeated $50$ times for each rotation and translation size. 

Note that~\cite{Rusinkiewicz:2019:ASO} estimates normals for the point clouds before sampling them. Hence, his registration results are based on information extracted from a much bigger point cloud. For a fair comparison, we perform two tests. In the first test, the full point cloud is exposed. In the second, only $1000$ points are given to~\cite{Rusinkiewicz:2019:ASO} for normal estimation.

Followed by~\cite{Rusinkiewicz:2019:ASO}, results are shown as a percentage of successful registration for different initial translation magnitudes and rotation angles. A success is defined if the aligned points are within a threshold ($1\%$ of point cloud size) of their ground-truth locations after alignment.

\noindent {\bf Full Point Clouds.}
We used transfer learning with the network trained on ModelNet40 (different sampling) as described in Section~\ref{different_sampling}. 
To adjust the network for the Stanford Bunny dataset~\cite{Turk:1994:ZPM}, we used a self-supervised learning technique for creating pairs of point clouds for training. 
In every iteration, two new sets of $1000$ points were sampled using farthest-point sampling from the same Bunny scan. A partial scan was simulated, as described in Section~\ref{modelnet40}. Then, a transformation, as described above, was applied and used for training the network. During evaluation, the results were fine tuned with Point-To-Plane ICP using the \textit{Open3D} python library \cite{Zhou2018}).

\begin{table}[t]
\begin{center}
  \includegraphics[width=\linewidth]{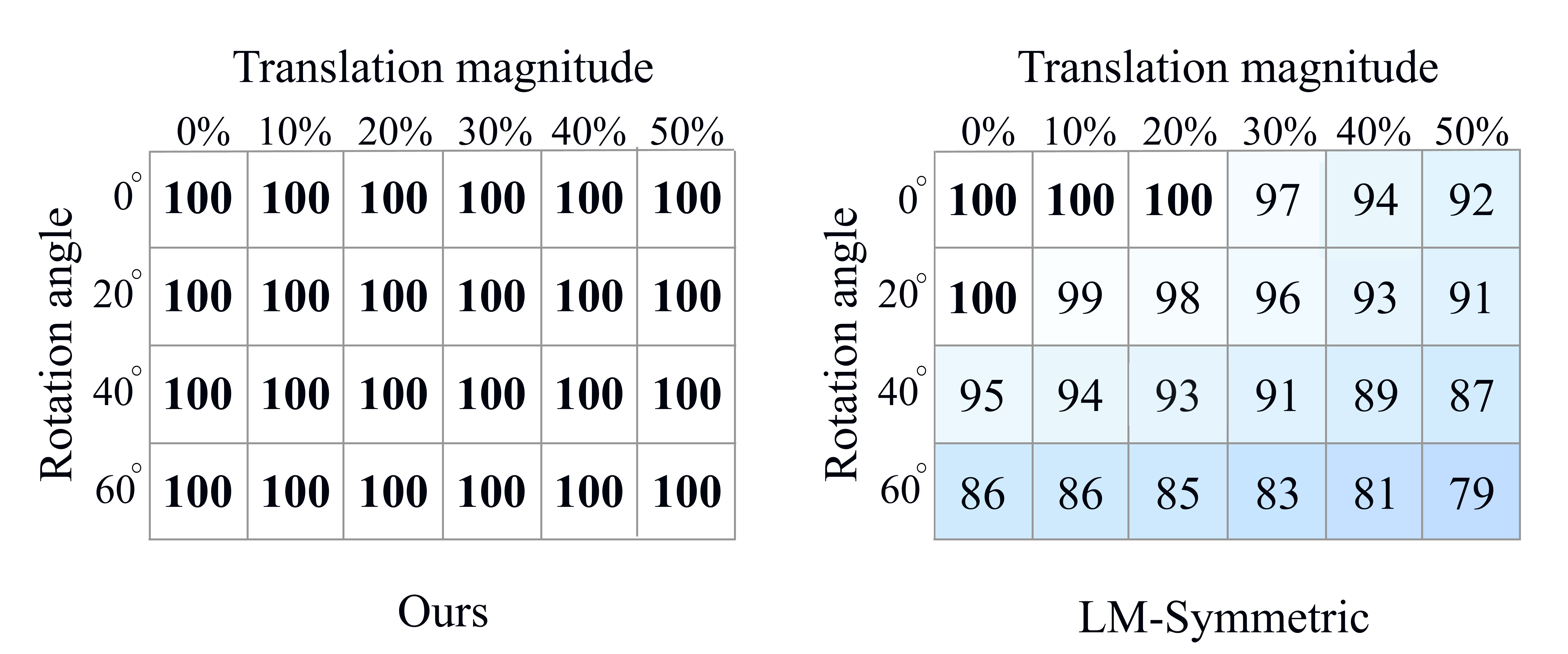}
\end{center}
\caption{Results of full point clouds of Bunny scans. {\bf Left:}  our results; {\bf Right:}  LM-Symmetric results~\cite{Rusinkiewicz:2019:ASO}. In this experiment, LM-Symmetric~\cite{Rusinkiewicz:2019:ASO} uses all of the scan points for normal estimation. Each cell in the table corresponds to a specific initial condition.} 
\label{table:bunny_full}
\end{table}

A comparison between \textbf{DeepBBS} and LM-Symmetric~\cite{Rusinkiewicz:2019:ASO} is shown in Table \ref{table:bunny_full}. \textbf{DeepBBS} performs better or equal in all initial transformation conditions, and achieves $100\%$ success. We also evaluated our technique with an error threshold of $0.5\%$ of point cloud size and still achieved $100\%$ success.


We report in the supplementary material (\ref{supp:ablation_bunny}) an ablation study for this experiment. It shows that removing the ICP fine-tuning hurts performance considerably. In addition, ICP by itself does not perform well. We also demonstrate the effect of farthest-point sampling.

\begin{table}[t]
\begin{center}
  \includegraphics[width=\linewidth]{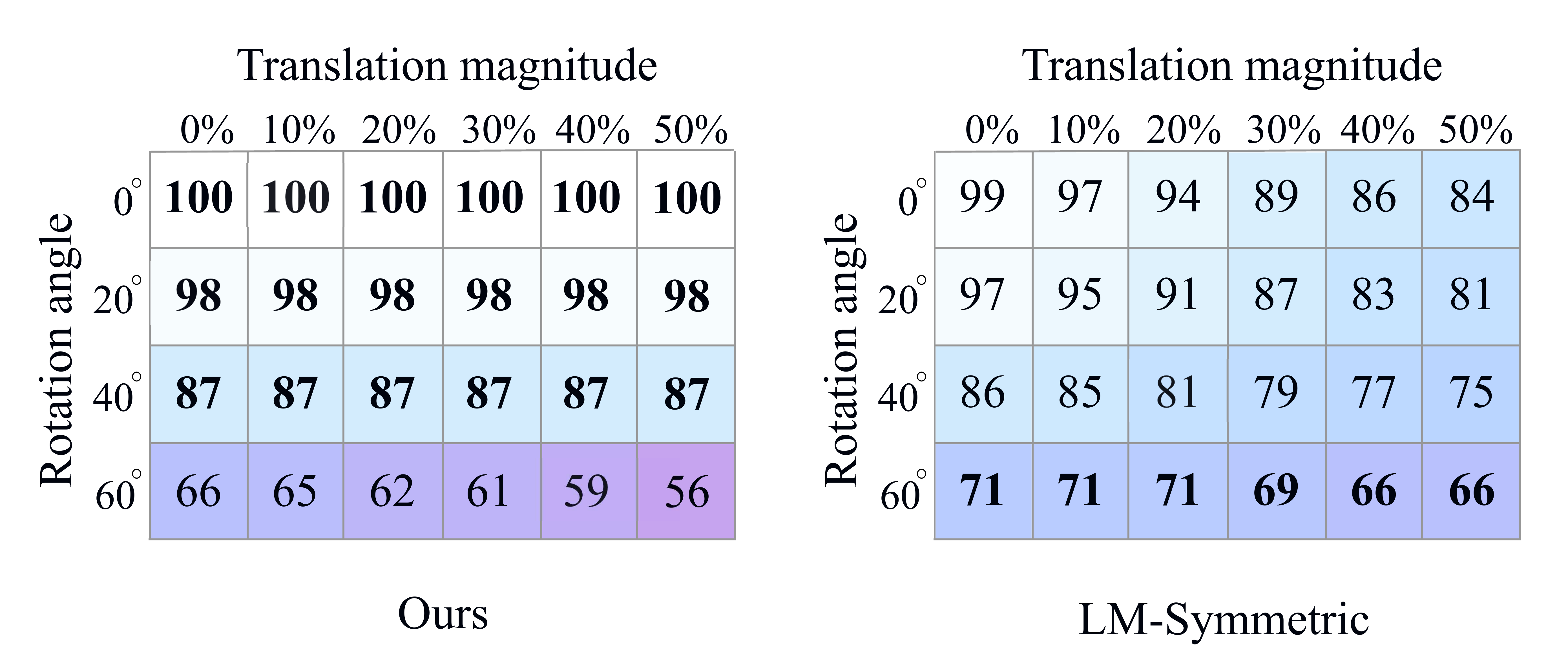}
\end{center}
\caption{Results of $1000$ points Bunny point clouds. {\bf Left:} Our results; {\bf Right:} LM-Symmetric results~\cite{Rusinkiewicz:2019:ASO}. In this experiment LM-Symmetric~\cite{Rusinkiewicz:2019:ASO} uses only $1000$ points for normal estimation. Each cell in the table corresponds to a specific initial condition.} 
\label{table:bunny_1000}
\end{table}

\begin{figure}
 \includegraphics[width=\linewidth]{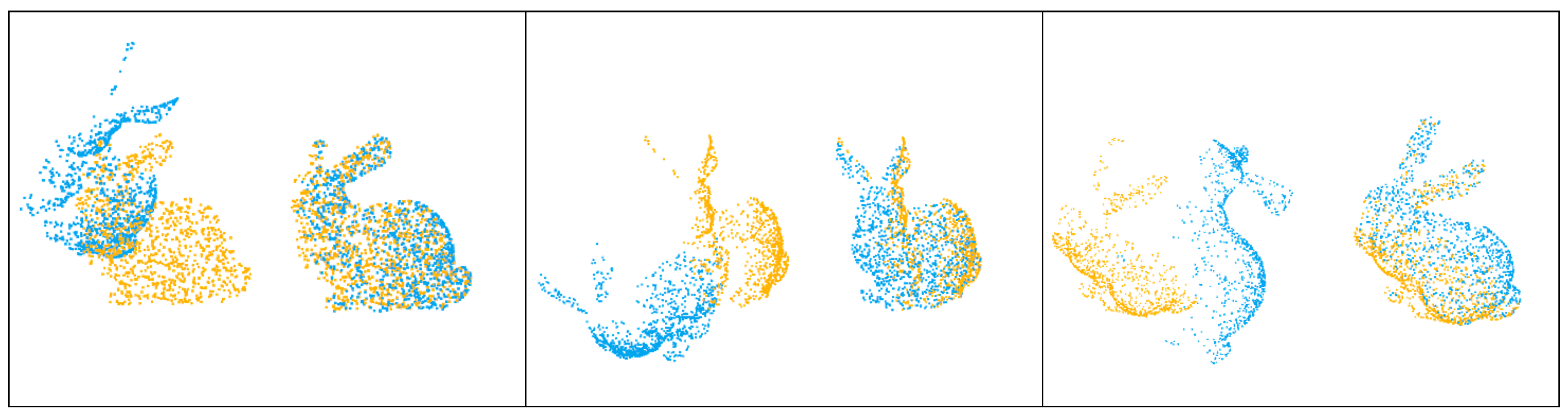}
\caption{Examples of point clouds from the Stanford bunny dataset before (left end of each frame) and after (right end of each frame) registration. The rotation angle is $60^{\circ}$, and the translation is $50\%$ of the shape size. These registrations are examples of the results of the experiment with $1000$ points in Section \ref{stanfordbunny}. Our method achieves accurate registration even for large rotations and translations and small overlap between scans.}
  \label{fig:bunny1000}
\end{figure}

\noindent {\bf $\mathbf{1000}$ Points Scans.} 
Here we use the same technique as before, except that we use the network that was trained on ModelNet40~\cite{ModelNet40} with noisy partial scans with its original weights.
Results are shown in Table \ref{table:bunny_1000}. \textbf{DeepBBS} performs better or equal in $18/24$ initial transformation conditions. These results demonstrate the ability of \textbf{DeepBBS} to generalize from one dataset to another. Examples of registrations in this experiment can be seen in Figure \ref{fig:bunny1000}.


\subsection{Apollo-SouthBay Dataset}
\label{apollo}
Apollo-SouthBay Dataset~\cite{apollo_dataset} contains 3D LiDAR scans. It covers different scenarios, including residential areas, urban downtown areas and highways. We followed DeepVCP's~\cite{DeepICP} benchmark. Frames were sampled at $100$ frames intervals with a maximal distance of $5m$ between them. Many of the sampled points are parts of the road and have less meaningful information for the registration. Therefore, we pre-processed the data with a road removal algorithm. Points within a  $0.5m$ layer around the least-squares plane fit were removed. Then, farthest-point sampling was applied to maintain $1000$ points.

We evaluate the results with the mean angular error (MAE), which is the Chordal distance~\cite{hartley2013rotation} between $\mathbf{R}$ and $\mathbf{R}_{GT}$, and with the mean transitional error (MTE), which is the Euclidean norm of the difference between $\mathbf{t}$ and $\mathbf{t}_{GT}$.

\begin{figure}
 \includegraphics[width=\linewidth]{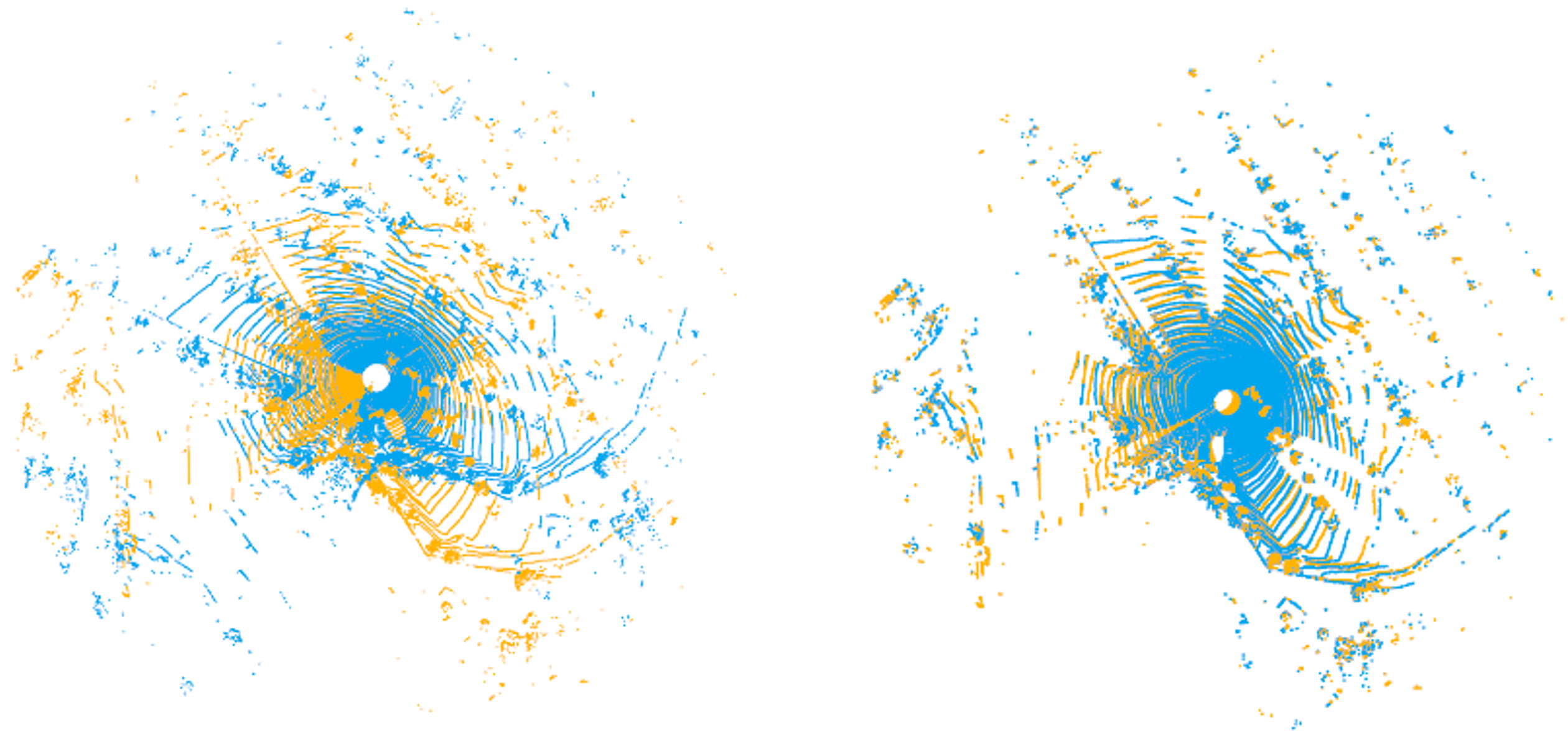}
\caption{Examples of point clouds from the Apollo-SouthBay dataset before (left) and after (right) \textbf{DeepBBS++} registration. Our method performs well on big scenes and large transformations.}
  \label{fig:apollo_example}
\end{figure}

A comparison between $\textbf{DeepBBS++}$, SymmetricICP~\cite{Rusinkiewicz:2019:ASO}, PRNet~\cite{PRNet} and ICP-Po2Po~\cite{ICP} is shown in Table \ref{table:apollo_comparisons}. We also tried to fine-tune our results with ICP instead of spatial BBS (Section \ref{inference}). The other methods were given point clouds without sampling as input. Our method showed significantly better results than the others. Spatial BBS fine-tuning was better than ICP fine-tunning. An example of our results is shown in Figure \ref{fig:apollo_example}.

Lu {\em et al}.~\cite{DeepICP} started the registration from an initial guess that is closer to the ground truth transformation. As our method has a wide basin of convergence, its results are similar with or without an initial guess. Comparison between the methods with an initial guess appears in \ref{supp:apollo_init_guess}.

\begin{table}
\begin{tabular}{|l|p{2cm}|p{2cm}|}
\hline
Method & MAE [$^{\circ}$] & MTE [$m$]\\
\hline\hline
SymmetricICP~\cite{Rusinkiewicz:2019:ASO} &     $0.58$  & $0.62$ \\
PRNet~\cite{PRNet}                        &     $1.24$     & $2.18$   \\  
ICP-Po2Po~\cite{ICP}                        &     $0.77$     & $0.64$   \\  
\hline
DeepBBS+ICP (ours) & $0.067$ & $0.058$ \\
DeepBBS++ (ours) & $\mathbf{0.058}$ & $\mathbf{0.046}$ \\
\hline
\end{tabular}
\caption{Results on the Apollo-SouthBay dataset. Our method outperforms all others.}
\label{table:apollo_comparisons}
\end{table}
\section{Conclusions}

This paper presented \textbf{DeepBBS}, a method for estimating the rigid transformation between two 3D point clouds. It is based on Best Buddies (i.e., mutual nearest neighbors), where a pair of points is said to be best buddies if one is the nearest neighbor of the other. Instead of finding best buddies in the input 3D space, we train a neural network to find an embedding space, in which the Best Buddies Similarity measure is computed. Experiments show that our method is robust to occlusions, has a very large basin of attractions, and achieves state-of-the-art results on several datasets.

{\small
\bibliographystyle{ieee_fullname}
\bibliography{for_Arxiv}

\begin{thebibliography}{10}\itemsep=-1pt

\bibitem{aberman2018neural}
Kfir Aberman, Jing Liao, Mingyi Shi, Dani Lischinski, Baoquan Chen, and Daniel
  Cohen-Or.
\newblock Neural best-buddies: Sparse cross-domain correspondence.
\newblock {\em ACM Transactions on Graphics (TOG)}, 37(4):1--14, 2018.

\bibitem{PointNetLK}
Yasuhiro Aoki, Hunter Goforth, Rangaprasad~Arun Srivatsan, and Simon Lucey.
\newblock Pointnetlk: Robust \& efficient point cloud registration using
  pointnet.
\newblock In {\em The IEEE Conference on Computer Vision and Pattern
  Recognition (CVPR)}, June 2019.

\bibitem{Bai_2021_CVPR}
Xuyang Bai, Zixin Luo, Lei Zhou, Hongkai Chen, Lei Li, Zeyu Hu, Hongbo Fu, and
  Chiew-Lan Tai.
\newblock Pointdsc: Robust point cloud registration using deep spatial
  consistency.
\newblock In {\em Proceedings of the IEEE/CVF Conference on Computer Vision and
  Pattern Recognition (CVPR)}, pages 15859--15869, June 2021.

\bibitem{bernard2020mina}
Florian Bernard, Zeeshan~Khan Suri, and Christian Theobalt.
\newblock Mina: Convex mixed-integer programming for non-rigid shape alignment.
\newblock In {\em Proceedings of the IEEE/CVF Conference on Computer Vision and
  Pattern Recognition}, pages 13826--13835, 2020.

\bibitem{ICP}
Paul~J. Besl and Neil~D. McKay.
\newblock A method for registration of 3-d shapes.
\newblock {\em IEEE Trans. Pattern Anal. Mach. Intell.}, 14(2):239--256, Feb.
  1992.

\bibitem{SparseICP}
Sofien Bouaziz, Andrea Tagliasacchi, and Mark Pauly.
\newblock Sparse iterative closest point.
\newblock In {\em Proceedings of the Eleventh Eurographics/ACMSIGGRAPH
  Symposium on Geometry Processing}, SGP '13, pages 113--123, 2013.

\bibitem{Chen:1992}
Yang Chen and G{\'e}rard Medioni.
\newblock Object modelling by registration of multiple range images.
\newblock {\em Image Vision Comput.}, 10(3):145--155, Apr. 1992.

\bibitem{TrimmedICP}
Dmitry Chetverikov, Dmitry Stepanov, and Pavel Krsek.
\newblock Robust euclidean alignment of 3d point sets: the trimmed iterative
  closest point algorithm.
\newblock {\em Image and Vision Computing}, 23(3):299 -- 309, 2005.

\bibitem{Choy_2020_CVPR}
Christopher Choy, Wei Dong, and Vladlen Koltun.
\newblock Deep global registration.
\newblock In {\em Proceedings of the IEEE/CVF Conference on Computer Vision and
  Pattern Recognition (CVPR)}, June 2020.

\bibitem{drory2020best}
Amnon Drory, Tal Shomer, Shai Avidan, and Raja Giryes.
\newblock Best buddies registration for point clouds, 2020.

\bibitem{Eckart_2018_ECCV}
B. Eckart, K. Kim, and J. Kautz.
\newblock Hgmr: Hierarchical gaussian mixtures for adaptive 3d registration.
\newblock In {\em The European Conference on Computer Vision (ECCV)}, September
  2018.

\bibitem{Fitzgibbon01c}
Andrew~W. Fitzgibbon.
\newblock Robust registration of {2D} and {3D} point sets.
\newblock In {\em British Machine Vision Conference}, pages 662--670, 2001.

\bibitem{FilterReg}
Wei Gao and Russ Tedrake.
\newblock Filterreg: Robust and efficient probabilistic point-set registration
  using gaussian filter and twist parameterization.
\newblock In {\em {IEEE} Conference on Computer Vision and Pattern Recognition,
  {CVPR}}, pages 11095--11104. Computer Vision Foundation / {IEEE}, 2019.

\bibitem{granger2002multi}
S{\'e}bastien Granger and Xavier Pennec.
\newblock Multi-scale em-icp: A fast and robust approach for surface
  registration.
\newblock In {\em European Conference on Computer Vision}, pages 418--432.
  Springer, 2002.

\bibitem{hartley2013rotation}
Richard Hartley, Jochen Trumpf, Yuchao Dai, and Hongdong Li.
\newblock Rotation averaging.
\newblock {\em International journal of computer vision}, 103(3):267--305,
  2013.

\bibitem{Hertz20PointGMM}
A. {Hertz}, R. {Hanocka}, R. {Giryes}, and D. {Cohen-Or}.
\newblock Pointgmm: A neural gmm network for point clouds.
\newblock In {\em IEEE/CVF Conference on Computer Vision and Pattern
  Recognition (CVPR)}, pages 12051--12060, 2020.

\bibitem{Jian:2011:RPS}
Bing Jian and Baba~C. Vemuri.
\newblock Robust point set registration using gaussian mixture models.
\newblock {\em IEEE Trans. Pattern Anal. Mach. Intell.}, 33(8):1633--1645, Aug.
  2011.

\bibitem{kingma2014adam}
Diederik~P Kingma and Jimmy Ba.
\newblock Adam: A method for stochastic optimization.
\newblock {\em arXiv preprint arXiv:1412.6980}, 2014.

\bibitem{DeepICP}
Weixin Lu, Guowei Wan, Yao Zhou, Xiangyu Fu, Pengfei Yuan, and Shiyu Song.
\newblock Deepvcp: An end-to-end deep neural network for point cloud
  registration.
\newblock {\em 2019 IEEE/CVF International Conference on Computer Vision
  (ICCV)}, Oct 2019.

\bibitem{apollo_dataset}
Weixin Lu, Yao Zhou, Guowei Wan, Shenhua Hou, and Shiyu Song.
\newblock L3-net: Towards learning based lidar localization for autonomous
  driving.
\newblock In {\em Proceedings of the IEEE Conference on Computer Vision and
  Pattern Recognition}, pages 6389--6398, 2019.

\bibitem{Lucas:1981}
Bruce~D. Lucas and Takeo Kanade.
\newblock An iterative image registration technique with an application to
  stereo vision.
\newblock In {\em Proceedings of the 7th International Joint Conference on
  Artificial Intelligence - Volume 2}, IJCAI'81, pages 674--679, 1981.

\bibitem{maron2016point}
Haggai Maron, Nadav Dym, Itay Kezurer, Shahar Kovalsky, and Yaron Lipman.
\newblock Point registration via efficient convex relaxation.
\newblock {\em ACM Transactions on Graphics (TOG)}, 35(4):1--12, 2016.

\bibitem{amo_fpcs_sig_08}
N.~J. Mitra, D. Aiger, and D. Cohen-Or.
\newblock 4-points congruent sets for robust surface registration.
\newblock {\em ACM Transactions on Graphics}, 27(3):\#85, 1--10, 2008.

\bibitem{myronenko2010point}
Andriy Myronenko and Xubo Song.
\newblock Point set registration: Coherent point drift.
\newblock {\em IEEE transactions on pattern analysis and machine intelligence},
  32(12):2262--2275, 2010.

\bibitem{BBS}
S. {Oron}, T. {Dekel}, T. {Xue}, W.~T. {Freeman}, and S. {Avidan}.
\newblock Best-buddies similarity—robust template matching using mutual
  nearest neighbors.
\newblock {\em IEEE Transactions on Pattern Analysis and Machine Intelligence},
  40(8):1799--1813, 2018.

\bibitem{pavlov2018aa}
Artem~L Pavlov, Grigory~WV Ovchinnikov, Dmitry~Yu Derbyshev, Dzmitry
  Tsetserukou, and Ivan~V Oseledets.
\newblock Aa-icp: Iterative closest point with anderson acceleration.
\newblock In {\em 2018 IEEE International Conference on Robotics and Automation
  (ICRA)}, pages 3407--3412. IEEE, 2018.

\bibitem{plotz2018neural}
Tobias Pl\"{o}tz and Stefan Roth.
\newblock {Neural Nearest Neighbors Networks}.
\newblock {\em Proceedings of Advances in Neural Information Processing Systems
  (NeuralIPS)}, 2018.

\bibitem{Pomerleau:2015}
F. {Pomerleau}, F. {Colas}, and R. {Siegwart}.
\newblock {\em A Review of Point Cloud Registration Algorithms for Mobile
  Robotics}.
\newblock now, 2015.

\bibitem{qi2017pointnet}
Charles~R. Qi, Hao Su, Kaichun Mo, and Leonidas~J. Guibas.
\newblock {PointNet: Deep Learning on Point Sets for 3D Classification and
  Segmentation}.
\newblock {\em Proceedings of the IEEE Conference on Computer Vision and
  Pattern Recognition (CVPR)}, pages 652--660, 2017.

\bibitem{Rusinkiewicz:2019:ASO}
Szymon Rusinkiewicz.
\newblock A symmetric objective function for {ICP}.
\newblock {\em ACM Transactions on Graphics (Proc. SIGGRAPH)}, 38(4), July
  2019.

\bibitem{RUS:2001}
S. {Rusinkiewicz} and M. {Levoy}.
\newblock Efficient variants of the icp algorithm.
\newblock In {\em Proceedings Third International Conference on 3-D Digital
  Imaging and Modeling}, pages 145--152, 2001.

\bibitem{FPFH}
Radu~Bogdan Rusu, Nico Blodow, and Michael Beetz.
\newblock Fast point feature histograms (fpfh) for 3d registration.
\newblock In {\em Proceedings of the 2009 IEEE International Conference on
  Robotics and Automation}, ICRA'09, page 1848–1853. IEEE Press, 2009.

\bibitem{Sarode2019PCRNetPC}
Vinit Sarode, Xueqian Li, Hunter Goforth, Yasuhiro Aoki, Rangaprasad~Arun
  Srivatsan, Simon Lucey, and Howie Choset.
\newblock Pcrnet: Point cloud registration network using pointnet encoding.
\newblock {\em ArXiv}, abs/1908.07906, 2019.

\bibitem{segal2009generalized}
Aleksandr Segal, Dirk Haehnel, and Sebastian Thrun.
\newblock Generalized-icp.
\newblock In {\em Robotics: science and systems}, volume~2, page 435. Seattle,
  WA, 2009.

\bibitem{SegalHT09}
Aleksandr Segal, Dirk Hähnel, and Sebastian Thrun.
\newblock Generalized-icp.
\newblock In Jeff Trinkle, Yoky Matsuoka, and José~A. Castellanos, editors,
  {\em Robotics: Science and Systems}. The MIT Press, 2009.

\bibitem{LSSVF}
Olga Sorkine-Hornung and Michael Rabinovich.
\newblock Least-squares rigid motion using svd.
\newblock {\em Technical note.}, 2016.

\bibitem{stoyanov2012fast}
Todor Stoyanov, Martin Magnusson, Henrik Andreasson, and Achim~J Lilienthal.
\newblock Fast and accurate scan registration through minimization of the
  distance between compact 3d ndt representations.
\newblock {\em The International Journal of Robotics Research},
  31(12):1377--1393, 2012.

\bibitem{tsin2004correlation}
Yanghai Tsin and Takeo Kanade.
\newblock A correlation-based approach to robust point set registration.
\newblock In {\em European conference on computer vision}, pages 558--569.
  Springer, 2004.

\bibitem{Turk:1994:ZPM}
Greg Turk and Marc Levoy.
\newblock Zippered polygon meshes from range images.
\newblock In {\em Proceedings of the 21st Annual Conference on Computer
  Graphics and Interactive Techniques}, SIGGRAPH '94, pages 311--318. ACM,
  1994.

\bibitem{Transformers}
Ashish Vaswani, Noam Shazeer, Niki Parmar, Jakob Uszkoreit, Llion Jones,
  Aidan~N Gomez, \L~ukasz Kaiser, and Illia Polosukhin.
\newblock Attention is all you need.
\newblock In I. Guyon, U.~V. Luxburg, S. Bengio, H. Wallach, R. Fergus, S.
  Vishwanathan, and R. Garnett, editors, {\em Advances in Neural Information
  Processing Systems 30}, pages 5998--6008. Curran Associates, Inc., 2017.

\bibitem{DCP}
Yue Wang and Justin~M. Solomon.
\newblock Deep closest point: Learning representations for point cloud
  registration.
\newblock In {\em The IEEE International Conference on Computer Vision (ICCV)},
  October 2019.

\bibitem{PRNet}
Yue Wang and Justin~M. Solomon.
\newblock Prnet: Self-supervised learning for partial-to-partial registration.
\newblock In Hanna~M. Wallach, Hugo Larochelle, Alina Beygelzimer, Florence
  d'Alch{\'{e}}{-}Buc, Emily~B. Fox, and Roman Garnett, editors, {\em Advances
  in Neural Information Processing Systems 32: Annual Conference on Neural
  Information Processing Systems 2019, NeurIPS 2019, 8-14 December 2019,
  Vancouver, BC, Canada}, pages 8812--8824, 2019.

\bibitem{DGCNN}
Yue Wang, Yongbin Sun, Ziwei Liu, Sanjay~E. Sarma, Michael~M. Bronstein, and
  Justin~M. Solomon.
\newblock Dynamic graph {CNN} for learning on point clouds.
\newblock {\em {ACM} Trans. Graph.}, 38(5):146:1--146:12, 2019.

\bibitem{Yang20TEASER}
Heng Yang, Jingnan Shi, and Luca Carlone.
\newblock Teaser: Fast and certifiable point cloud registration.
\newblock {\em IEEE Transactions on Robotics}, 2020.

\bibitem{yew20183dfeat}
Zi~Jian Yew and Gim~Hee Lee.
\newblock {3DFeat-Net}: Weakly supervised local 3d features for point cloud
  registration.
\newblock In {\em European Conference on Computer Vision}, pages 630--646.
  Springer, 2018.

\bibitem{yew2020-RPMNet}
Zi~Jian Yew and Gim~Hee Lee.
\newblock Rpm-net: Robust point matching using learned features.
\newblock In {\em Conference on Computer Vision and Pattern Recognition
  (CVPR)}, 2020.

\bibitem{yuan2020deepgmr}
Wentao Yuan, Benjamin Eckart, Kihwan Kim, Varun Jampani, Dieter Fox, and Jan
  Kautz.
\newblock Deepgmr: Learning latent gaussian mixture models for registration.
\newblock In {\em ECCV}, 2020.

\bibitem{ModelNet40}
{Zhirong Wu}, S. {Song}, A. {Khosla}, {Fisher Yu}, {Linguang Zhang}, {Xiaoou
  Tang}, and J. {Xiao}.
\newblock 3d shapenets: A deep representation for volumetric shapes.
\newblock In {\em 2015 IEEE Conference on Computer Vision and Pattern
  Recognition (CVPR)}, pages 1912--1920, 2015.

\bibitem{FGR}
Qian{-}Yi Zhou, Jaesik Park, and Vladlen Koltun.
\newblock Fast global registration.
\newblock In Bastian Leibe, Jiri Matas, Nicu Sebe, and Max Welling, editors,
  {\em Computer Vision - {ECCV} 2016 - 14th European Conference, Amsterdam, The
  Netherlands, October 11-14, 2016, Proceedings, Part {II}}, volume 9906 of
  {\em Lecture Notes in Computer Science}, pages 766--782. Springer, 2016.

\bibitem{Zhou2018}
Qian-Yi Zhou, Jaesik Park, and Vladlen Koltun.
\newblock {Open3D}: {A} modern library for {3D} data processing.
\newblock {\em arXiv:1801.09847}, 2018.

\end{thebibliography}
}
\clearpage
\onecolumn
\appendixpageoff
\appendixtitleoff
\begin{appendices}
  \setcounter{section}{19}
  \crefalias{section}{supp}
  \section*{}

{\LARGE\bfseries  Supplementary Material}
\vspace{1cm}

\thispagestyle{empty}

The supplementary material consists of a number of subjects. First, Complementary figures that demonstrate our method for the method chapter in the paper. Second, An ablation study for two of the tests that were performed. The first test is of partial scans with unseen point clouds on ModelNet40~\cite{ModelNet40} and the second test is of real scans of the Stanford Bunny~\cite{Turk:1994:ZPM} with full point clouds. Then we add more comparison results with other methods on the ModelNet40~\cite{ModelNet40} tests. Next, Results from the Apollo-SouthBay~\cite{apollo_dataset} dataset with an initial guess. Finally, we supply more visual examples for our results and a visual comparison to other techniques.

\subsection{More Figures for Illustrating the Method}
\label{supp:more_figures}
\begin{figure}[!h]
\begin{center}
  \includegraphics[width=10cm]{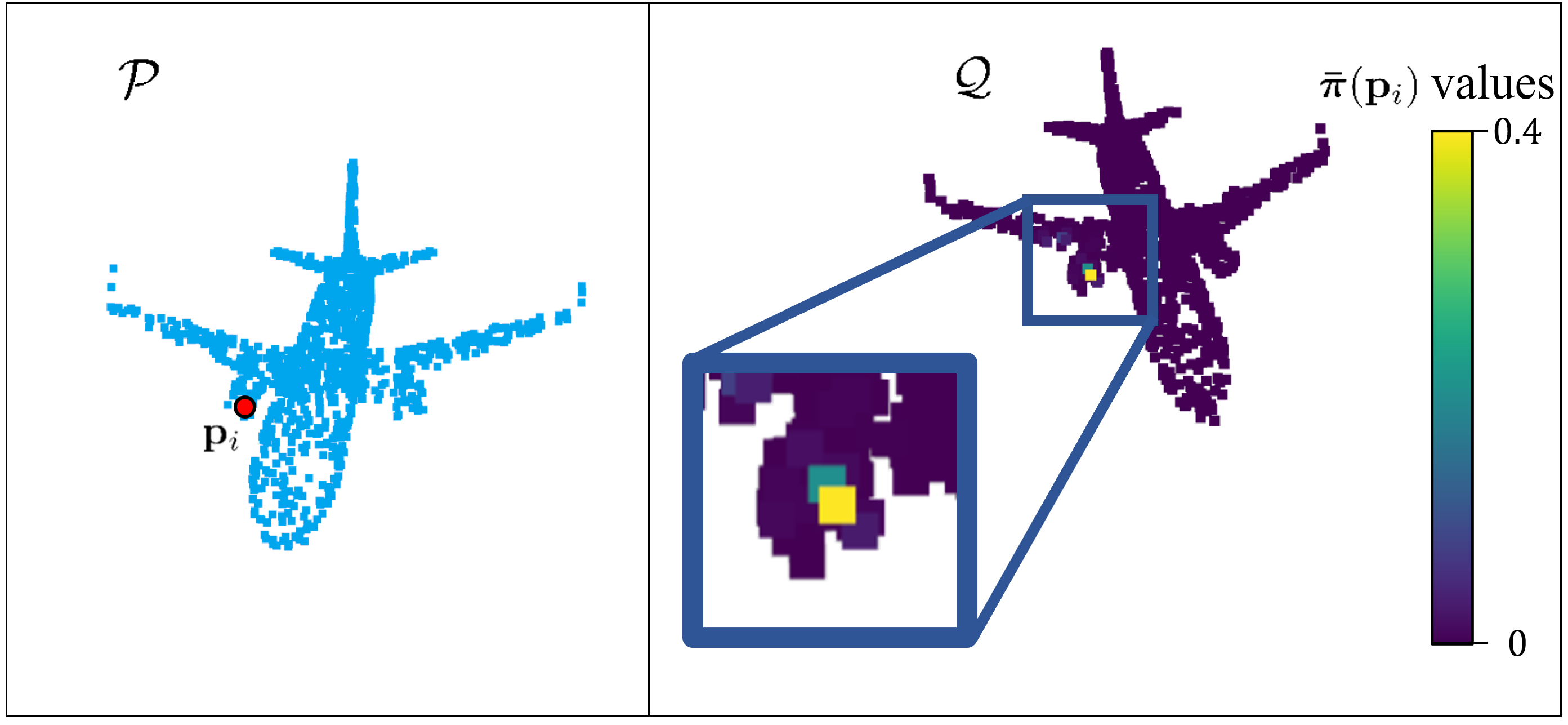}
\caption{\emph{Illustration of creating $\mathcal{\hat Q}$.}  \textbf{Left:} Point cloud $\mathcal{P}$. A particular point $\mathbf{p}_i$ is marked with a red circle. \textbf{Right:} point cloud $\mathcal{Q}$. The points in the point cloud are coloured with $\boldsymbol{\bar\pi}(\mathbf{p}_i)$. In the enlarged area it can be noticed that there are two points in the correlative part of the airplane, which have a significant $\bar\pi$ value. Hence, they will be the dominant points of the weighted sum for calculating $\mathbf{{\hat q}_i}$ in Equation~(\ref{eq:calc_q_hat}).}
  \label{fig:qhat}
\end{center}
\end{figure}

\begin{figure}[!h]
\begin{center}
  \includegraphics[width=10cm]{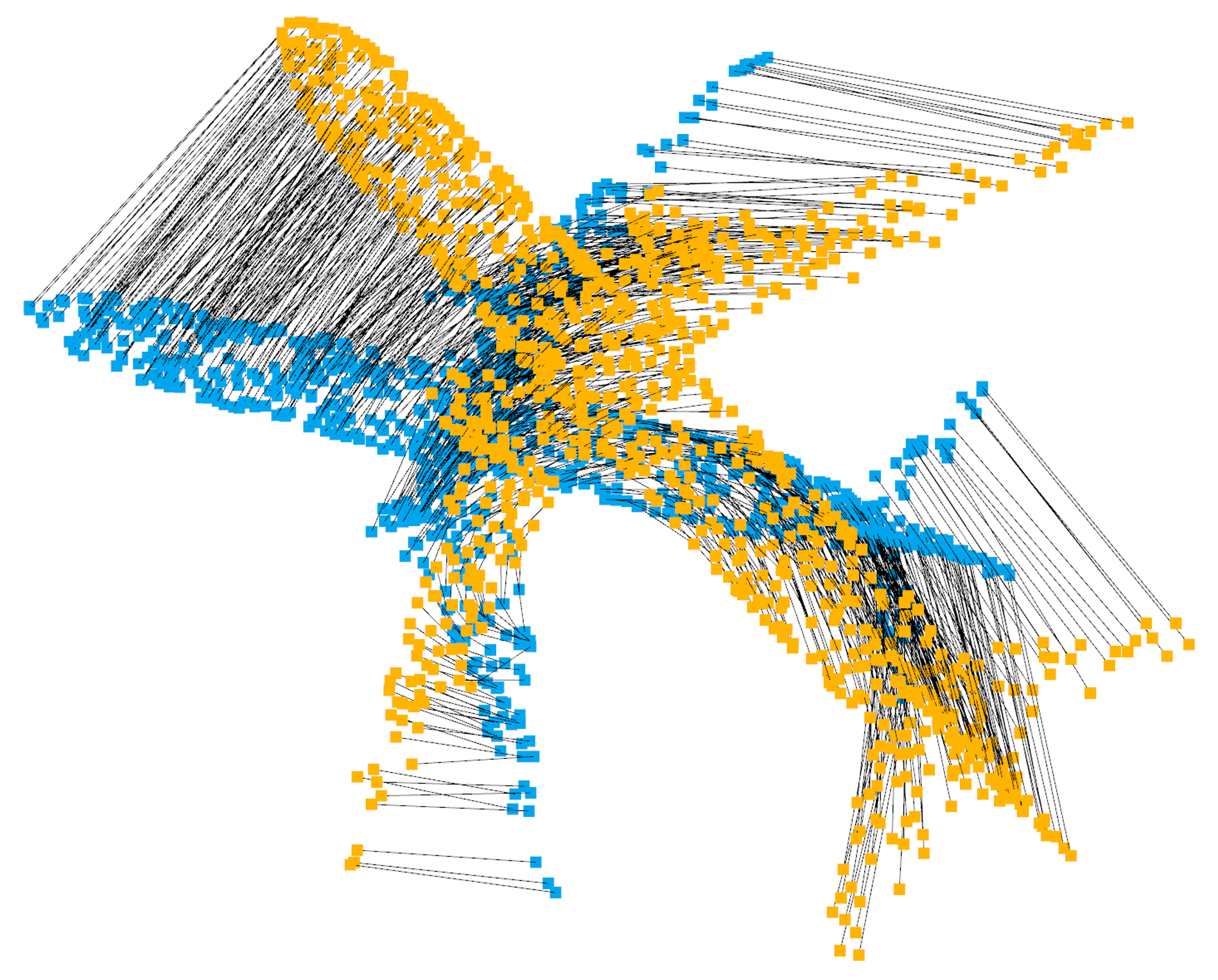}
\caption{Corresponding points of $\mathcal{P}$ and $\mathcal{\hat Q}$. $\mathcal{P}$ is coloured with blue and $\mathcal{\hat Q}$ is coloured with orange. Between points in each pair $(\mathbf{p}_i, \mathbf{{\hat q}}_i)$ there is a line indicating a correspondence. }
  \label{fig:p_qhat_corr}
\end{center}
\end{figure}

In this section we present two figures that illustrate our method, as was describes in Section~\ref{method}. Figure \ref{fig:qhat} shows an illustration of the process of creating $\mathcal{\hat Q}$ (referenced from Section~\ref{mapping}). Figure \ref{fig:p_qhat_corr} shows an example for correspondences between points in $\mathcal{P}$ and $\mathcal{\hat Q}$. It demonstrates the semantic connection between points of the same part of the object (referenced from Section~\ref{mapping}).

\subsection{Ablation Study}
\subsubsection{Partial Scans - Unseen Point Clouds}
\label{supp:ablation_modelnet40}
\label{ablation}

Table~\ref{table:ablation} contains the performed ablation tests. We show the importance of every element in our method by eliminating it and then reevaluating the method. The tests follow the settings used in the experiment of partial scans with unseen point clouds. The results indicate that every component in our method improves the results. Specifically, the partial reevaluated methods result in complete failure in two cases. The first case is when only spatial fine-tuning is applied (without passing through the network first). The failure occurs because spatial fine-tuning works best when the point clouds are close to each other, and thus similarity in the 3D space is significant. Without the network bringing the point clouds close to each other, especially from a distant initial position, the spatial fine-tuning fails to converge to accurate results. The second case in which the method fails is when $\gamma$ is omitted. This happens because without $\gamma$, pairs of points are not filtered or weighted. Points without matching points in the other point cloud still affect the registration, which leads to poor results.

A short explanation of the preformed test:
\begin{enumerate}
    \item \textbf{w/o spatial fine-tuning}- fine-tuning in the 3D space (described in Section~\ref{inference}) was eliminated during inference. We call this algorithm (without fine-tuning) DeepBBS.
    \item \textbf{Only spatial fine-tuning-} we tested performances using just the spatial fine-tuning step without the previous network step. This test shows that spatial fine-tuning improves results as fine-tuning is not sufficient by itself.
    \item \textbf{w/o iterations-} iterative passing through the network was dismissed, i.e., before fine-tuning, only one iteration of passing through the network was performed.
    \item \textbf{w/o decreasing T-} The temperature parameter, $T$,  which balances the similarity in the feature space and the 3D input space, was not decreased in every iteration.
    \item \textbf{w/o $\gamma$-} the correspondences' weights, $\gamma$, were eliminated (by setting $\gamma_i=1, \forall i$) during training and evaluation.
    \item \textbf{w/o spatial part of $\gamma$-} the spatial part in the definition of $\gamma$ (Equation~(\ref{eq:gamma_calc})) was eliminated during training and evaluation. (i.e. $\gamma_i = \sum_{j=1}^{M} {{\tilde B}_{ij}}$ instead of $\gamma_i = \sum_{j=1}^{M} {{\tilde B}_{ij}}e^{-D_{ij}/T}$).
    \item \textbf{w/o pointwise loss decays-} we tested the effect of the decay of the third summand in the definition of the loss (Equation (\ref{eq:loss})) by setting $\beta = 1$ during training. This prevents the term from decaying.
    \item \textbf{w/o pointwise loss-} we tested the third summand's influence in the definition of the loss function by setting $\beta = 0$ during training.
\end{enumerate}

\begin{table}[h!]
\begin{center}
\begin{tabular}{|l|c|c|c|c|c|c|}
\hline
Method & MSE($\mathbf{R}$) & RMSE($\mathbf{R}$) & MAE($\mathbf{R}$) &
         MSE($\mathbf{t}$) & RMSE($\mathbf{t}$) & MAE($\mathbf{t}$) \\
\hline\hline
\textbf{DeepBBS++ - Full} & $\mathbf{0.0002}$ & $\mathbf{0.014}$& $\mathbf{0.006}$  & $\mathbf{0.0000001}$ & $\mathbf{0.0004}$ & $\mathbf{0.0001}$ \\
\hline
w/o spatial fine-tuning    & $0.002$   & $0.041$  & $0.021$  & $0.000001$ & $0.0007$ & $0.0004$ \\
only spatial fine-tuning & $535.398$   & $23.139$  & $14.189$  & $0.023$ & $0.152$ & $0.110$ \\
w/o iterations    & $1.253$   & $1.119$  & $0.151$  & $0.0003$ & $0.017$ & $0.004$ \\
w/o decreasing $T$  & $0.0002$   & $0.014$  & $0.006$  & $0.0000006$ & $0.0008$ & $0.0002$ \\  
w/o $\gamma$ & $48.853$ & $6.990$ & $2.922$ & $0.012$ & $0.108$ & $0.067$ \\
w/o spatial part of $\gamma$ & $0.009$ & $0.097$ & $0.009$ & $0.00001$ & $0.003$ & $0.0003$ \\
w/o pointwise loss decays & $1.675$ & $1.294$ & $0.058$ & $0.0002$ & $0.013$ & $0.001$\\
w/o pointwise loss& $2.086$ & $1.444$ & $0.089$ & $0.0002$ & $0.013$ & $0.001$\\
\hline
\end{tabular}
\end{center}
\caption{\emph{Ablation study for ModelNet40 - unseen point clouds experiment.} The first row shows our complete scheme. The rest of the rows show our scheme without one element.}
\label{table:ablation}
\end{table}

\subsubsection{Stanford Bunny - Full Point Clouds}
\label{supp:ablation_bunny}
In Table~\ref{table:ablation_bunny}, an ablation test for the real scans experiment of full point clouds of the Stanford Bunny dataset~\cite{Turk:1994:ZPM} is shown. As can be seen, eliminating each component from the method causes significant performance degradation. From the results, it can be concluded that the network brings the point clouds into the ICP's basin of convergence, which fine-tunes the result into an exact registration. Another conclusion is that farthest-point sampling improves the results in comparison to random sampling.

\begin{table}[h!]
\begin{center}
 \includegraphics[width=\linewidth]{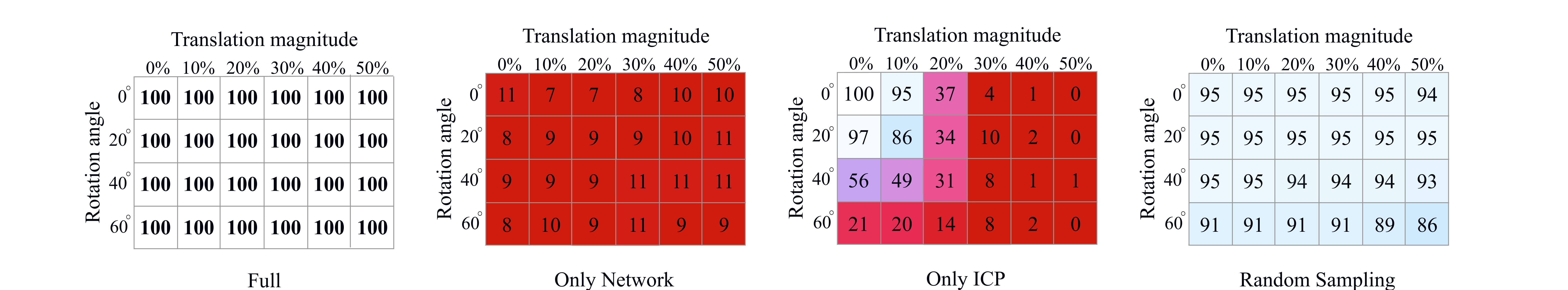}
 \end{center}
\caption{\emph{Ablation tests for Stanford Bunny dataset with full point clouds.} The left table shows the complete method, which is composed of a network and ICP as fine-tuning and farthest-point sampling. Each cell in the tables corresponds to a specific initial condition.}
  \label{table:ablation_bunny}
\end{table}

\subsection{More comparisons in the ModelNet40 tests}
\label{supp:more_comparisons}
Tables~\ref{table:extension_table_unseen_pc},~\ref{table:extension_table_unseen_categories},~\ref{table:extension_table_gaussian_noise} and~\ref{table:extension_table_different_samplings} include more comparisons for the experiments presented in Tables~\ref{table:unseen_point_clouds},~\ref{table:unseen_categories},~\ref{table:noise} and~\ref{table:bbs_comparisons} correspondingly. The additional methods are BBR, BD and BDN~\cite{drory2020best}, which are classic methods for point cloud registration that use the best buddies similarity measure. Another method is 4PCS~\cite{amo_fpcs_sig_08} which is a RANSAC-based method that does not rely on the initial conditions. Table~\ref{table:extension_table_different_samplings} does not include results on BBR, BD and BDN because these results are presented in Table~\ref{table:bbs_comparisons}.

\begin{table}[h!]
\begin{center}
\begin{tabular}{|l|c|c|c|c|c|c|}
\hline
Method & MSE($\mathbf{R}$) & RMSE($\mathbf{R}$) & MAE($\mathbf{R}$) &
         MSE($\mathbf{t}$) & RMSE($\mathbf{t}$) & MAE($\mathbf{t}$) \\
\hline\hline

BBR~\cite{drory2020best}  &  $1115.843$ & $33.404$ & $22.027$ & $0.0521$ & $0.2283$ & $0.1829$  \\
BD~\cite{drory2020best}  &  $872.973$ & $29.546$ & $24.975$ & $0.0824$ & $0.2870$ & $0.2485$ \\
BDN~\cite{drory2020best} &  $672.381$ & $25.930$ & $22.405$ & $0.0825$ & $0.2873$ & $0.2488$ \\
4PCS~\cite{amo_fpcs_sig_08}      & $2822.730$  & $53.129$ & $21.227$  & $0.0038$ & $0.061$ & $0.029$ \\

\hline
DeepBBS (ours) & $0.002$ & $0.041$& $0.021$  & $0.000001$ & $0.0007$ & $0.0004$ \\
DeepBBS++ (ours) & $\mathbf{0.0002}$ & $\mathbf{0.014}$& $\mathbf{0.006}$  & $\mathbf{0.0000001}$ & $\mathbf{0.0004}$ & $\mathbf{0.0001}$ \\
\hline
\end{tabular}
\end{center}
\caption{More results on unseen point clouds (extension to Table~\ref{table:unseen_point_clouds}). Best result in each criterion among all of the test methods (in Section~\ref{modelnet40} and in the sup. mat.) are marked in bold.}
\label{table:extension_table_unseen_pc}

\end{table}

\begin{table}[h!]
\begin{center}
\begin{tabular}{|l|c|c|c|c|c|c|}
\hline
Method & MSE($\mathbf{R}$) & RMSE($\mathbf{R}$) & MAE($\mathbf{R}$) &
         MSE($\mathbf{t}$) & RMSE($\mathbf{t}$) & MAE($\mathbf{t}$) \\
\hline\hline

BBR~\cite{drory2020best}  & $1071.025$  &   $32.727$ & $21.615$ & $0.0518$ & $0.2275$ & $0.1827$ \\ 
BD~\cite{drory2020best}  &  $830.947$   &   $28.826$ & $24.483$ & $0.0830$ & $0.2881$ & $0.2501$ \\ 
BDN~\cite{drory2020best} &  $675.065$   &   $25.982$ & $22.481$ & $0.0830$ & $0.2881$ & $0.2502$ \\ 
4PCS~\cite{amo_fpcs_sig_08} & $2690.190$  & $51.867$ & $20.053$  & $0.0043$ & $0.066$ & $0.031$ \\

\hline
DeepBBS (ours) & $0.006$ & $0.075$& $0.040$  & $0.000001$ & $0.0011$ & $0.0006$ \\
DeepBBS++ (ours) & $\mathbf{0.0006}$ & $\mathbf{0.024}$& $\mathbf{0.008}$  & $\mathbf{0.0000002}$ & $\mathbf{0.0005}$ & $\mathbf{0.0002}$ \\
\hline
\end{tabular}
\end{center}
\caption{More results on unseen categories (extension to Table~\ref{table:unseen_categories}). Best result in each criterion among all of the test methods (in Section~\ref{modelnet40} and in the sup. mat.) are marked in bold.}
\label{table:extension_table_unseen_categories}
\end{table}

\begin{table}[h!]
\begin{center}
\begin{tabular}{|l|c|c|c|c|c|c|}
\hline
Method & MSE($\mathbf{R}$) & RMSE($\mathbf{R}$) & MAE($\mathbf{R}$) &
         MSE($\mathbf{t}$) & RMSE($\mathbf{t}$) & MAE($\mathbf{t}$) \\
\hline\hline

BBR~\cite{drory2020best}  &  $1063.096$ &  $32.605$ & $21.528$ & $0.0515$ & $0.227$ &  $0.1816$ \\
BD~\cite{drory2020best}  &   $874.637 $ &  $29.574$ & $25.024$ & $0.0824$ & $0.287$ &  $0.2486$ \\ 
BDN~\cite{drory2020best} &   $672.453 $ &  $25.932$ & $22.423$ & $0.0825$ & $0.287$ &  $0.2487$ \\ 
4PCS~\cite{amo_fpcs_sig_08} & $3643.962$  & $60.365$ & $27.179$  & $0.0046$ & $0.068$ & $0.034$ \\

\hline
DeepBBS (ours)     & $17.625$   & $4.198$  & $1.715$  & $0.0019$ & $0.0441$ & $0.023$ \\
DeepBBS++ (ours)     & $\mathbf{16.568}$   & $\mathbf{4.070}$  & $\mathbf{0.974}$  & $0.0022$ & $0.0471$ & $0.017$ \\
\hline
\end{tabular}
\end{center}
\caption{More results on corrupted with white Gaussian noise (extension to Table~\ref{table:noise}). Best result in each criterion among all of the test methods (in Section~\ref{modelnet40} and in the sup. mat.) are marked in bold.}
\label{table:extension_table_gaussian_noise}
\end{table}

\begin{table}[h!]
\begin{center}
\begin{tabular}{|l|c|c|c|c|c|c|}
\hline
Method & MSE($\mathbf{R}$) & RMSE($\mathbf{R}$) & MAE($\mathbf{R}$) &
         MSE($\mathbf{t}$) & RMSE($\mathbf{t}$) & MAE($\mathbf{t}$) \\
\hline\hline

4PCS~\cite{amo_fpcs_sig_08}      & $6014.433$  & $77.553$ & $43.614$  & $0.001947$ & $0.044$ & $0.029$ \\

\hline
DeepBBS (ours) & $\mathbf{8.566}$ & $\mathbf{2.927}$& $\mathbf{1.089}$  & $\mathbf{0.00006}$ & $\mathbf{0.008}$ & $\mathbf{0.006}$ \\
DeepBBS++ (ours) & $11.914$ & $3.452$& $1.640$  & $0.00044$ & $0.021$ & $0.016$ \\
\hline
\end{tabular}
\end{center}
\caption{More results on point clouds of different samplings (extension to Table~\ref{table:bbs_comparisons}). Best result in each criterion among all of the test methods (in Section~\ref{modelnet40} and in the sup. mat.) are marked in bold.}
\label{table:extension_table_different_samplings}
\end{table}

\subsection{Apollo-SouthBay with initial guess}
\label{supp:apollo_init_guess}
Lu {\em et al.}~\cite{DeepICP} evaluated their method using an initial guess for the transformation between the point clouds. A uniformly distributed random error in the range of $0m - 1m$ in the $x-y-z$ dimensions was added to the point clouds when they were aligned with the ground truth transformation, and a random error in the range of $0^{\circ}-1^{\circ}$ was added in the $roll-pitch-yaw$ dimensions.

The results in Section~\ref{apollo} do not include an initial transformation. Table \ref{table:with_init_guess} contains a comparison of our methods to other techniques with an initial transformation, as was reported in \cite{DeepICP}. We report results of DeepBBS that was fine-tuned with ICP and of DeepBBS++.

DeepBBS++'s performances are similar to those reported in Table~\ref{table:apollo_comparisons}, where no initial guess was given. We conclude that the basin of convergence of DeepBBS++ is wide because an initial guess does not improve the results. DeepBBS++ bases its registration on a relatively small amount of points ($1000$ in this experiment) and still poses as a competitor to other methods that use the entire point cloud.

\begin{table}[h!]
\begin{center}
\begin{tabular}{|l|c|c|}
\hline
Method & Mean Angular Error [$^{\circ}$] & Mean Transitional Error [$m$]\\
\hline\hline
ICP-Po2Po~\cite{ICP} &     $0.051$     & $0.089$   \\  
ICP-Po2Pl~\cite{ICP} &     $0.026$     & $0.024$   \\  
G-ICP~\cite{segal2009generalized} &     $\mathbf{0.025}$     & $\mathbf{0.014}$   \\  
AA-ICP~\cite{pavlov2018aa} &     $0.054$     & $0.109$   \\
NDT-P2D~\cite{stoyanov2012fast} &     $0.045$     & $0.045$   \\  
CPD~\cite{myronenko2010point} &     $0.054$     & $0.210$   \\  
3DFeat-Net~\cite{yew20183dfeat} &     $0.076$     & $0.061$   \\  
DeepICP~\cite{DeepICP} &     $0.056$     & $0.018$   \\  
\hline
DeepBBS + ICP (ours) & $0.067$ & $0.058$ \\
DeepBBS++ (ours) & $0.059$ & $0.047$ \\
\hline
\end{tabular}
\end{center}
\caption{Results on Apollo-SouthBay dataset~\cite{apollo_dataset} with an initial guess.}
\label{table:with_init_guess}
\end{table}

\subsection{More visual examples}
\label{supp:more_visual_exp}
\subsubsection{Registration Results of different categories}
\begin{figure}[h]
 \includegraphics[width=\linewidth]{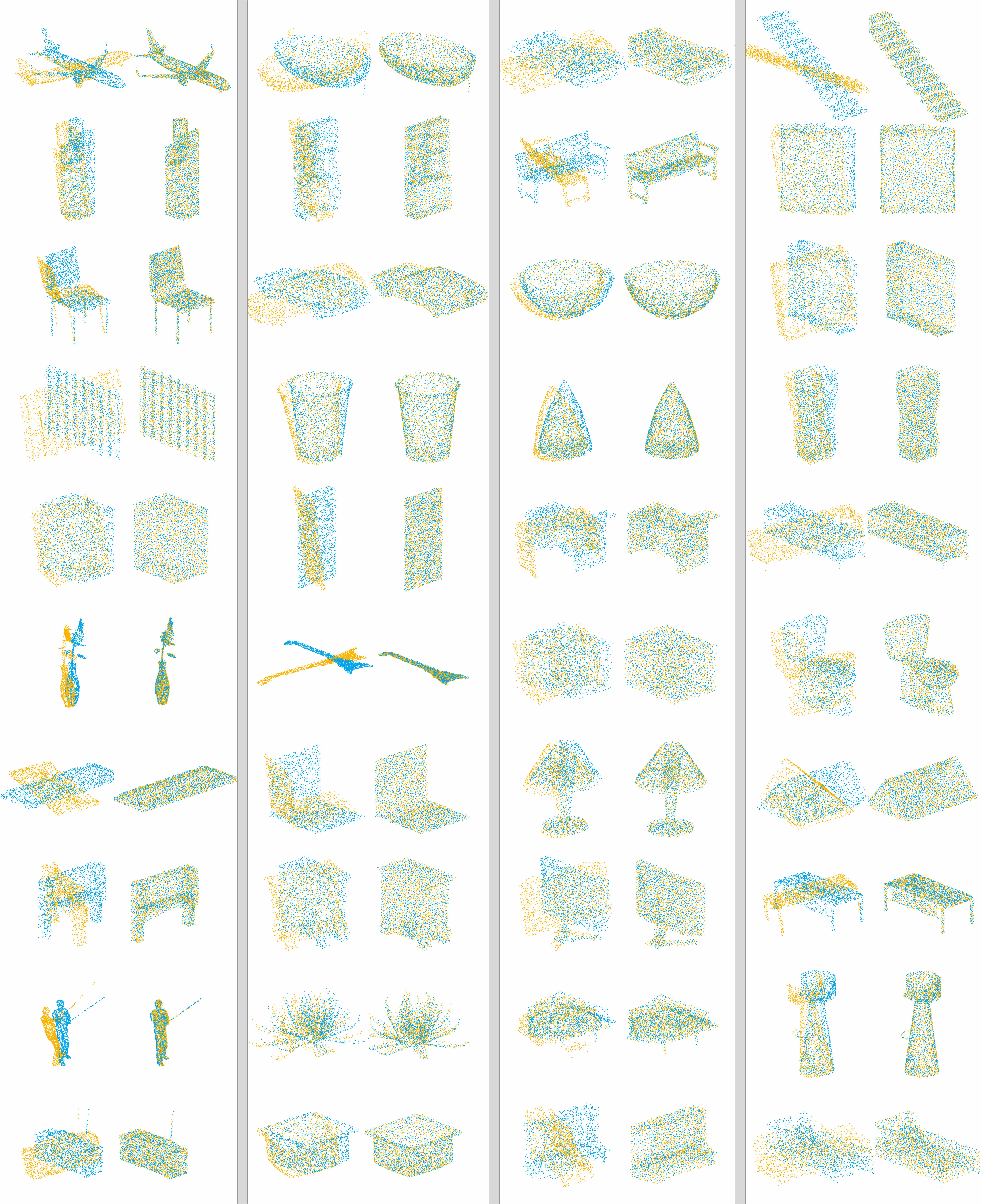}
\caption{\emph{Examples of registration results from different samplings test.} The first model of each category of ModelNet40~\cite{ModelNet40} dataset is shown before (left) and after registration (right).}
  \label{fig:examples}
\end{figure}

\begin{figure}[h]
 \includegraphics[width=\linewidth]{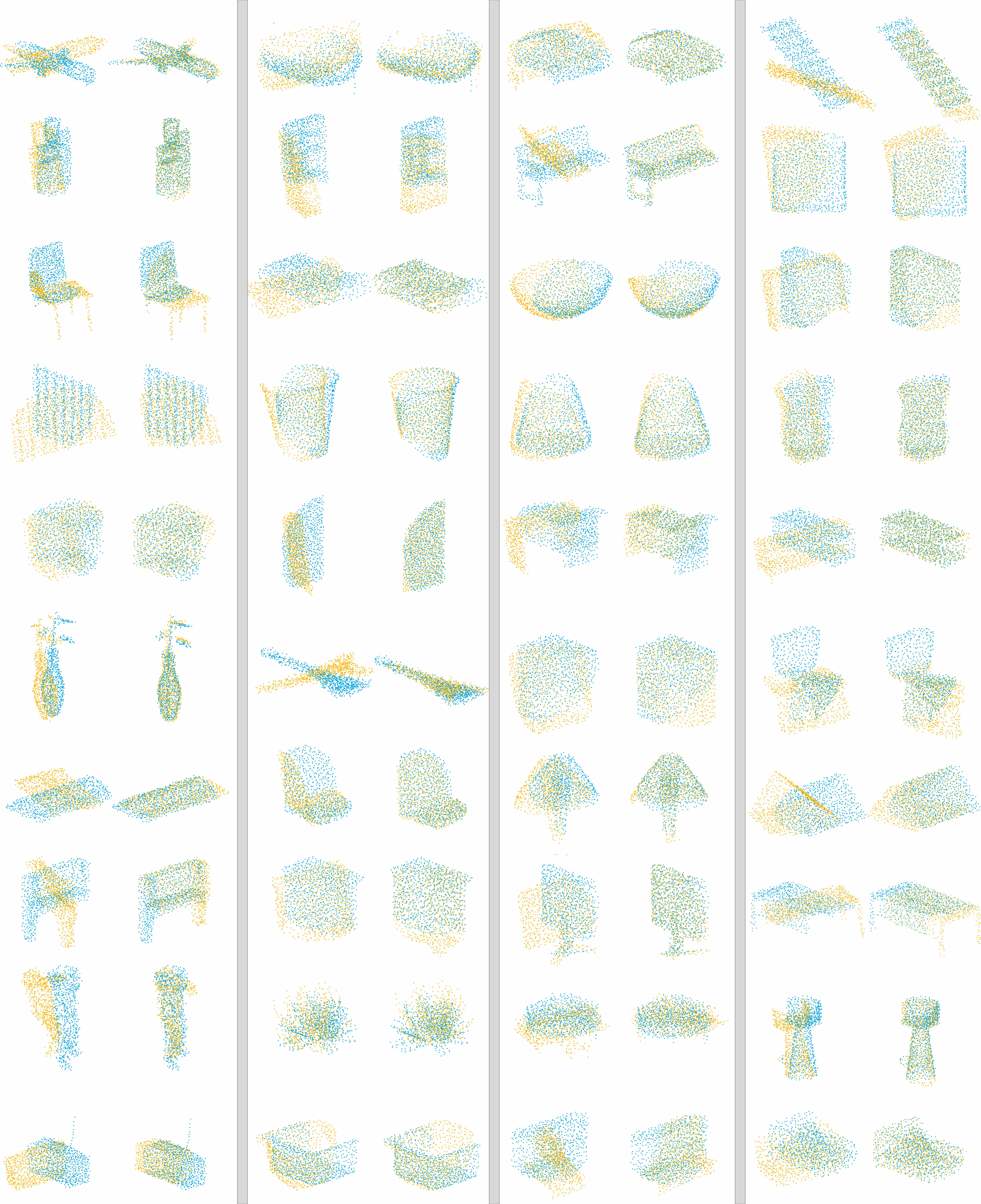}
\caption{\emph{Examples of registration results from partial scans with Gaussian noise test.} The first model of each category of ModelNet40~\cite{ModelNet40} dataset is shown before (left) and after registration (right).}
  \label{fig:examples_noise}
\end{figure}

The next figures show registration examples of the ModelNet40~\cite{ModelNet40} dataset. The shown tests are of different point samplings and of partial scans with Gaussian noise in Figures~\ref{fig:examples} and~\ref{fig:examples_noise} respectively. The presented results were not chosen for their quality, but for being the first pair of point clouds in each category of ModelNet40~\cite{ModelNet40}.

\subsubsection{Comparison to other methods}
\begin{figure}[h!]
 \includegraphics[width=\linewidth]{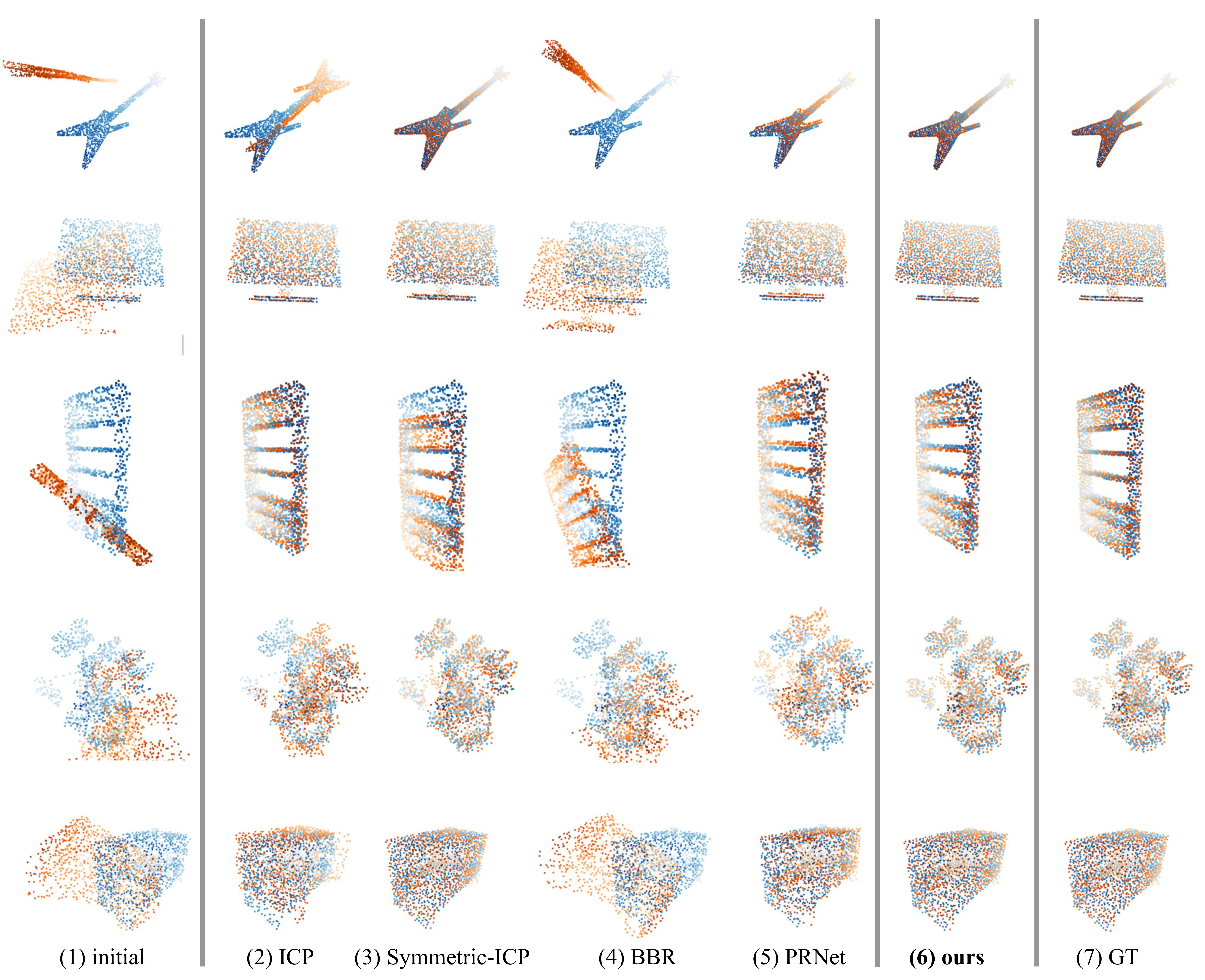}
\caption{\emph{Comparison with other methods - different sampling test.} Each row contains different pairs of point clouds from the ModelNet40 dataset~\cite{ModelNet40}; (1) initial pose of the point clouds before registration; (2) ICP \cite{ICP}; (3) Symmetric-ICP~\cite{Rusinkiewicz:2019:ASO}; (4) BBR~\cite{drory2020best}; (5) PRNet~\cite{PRNet}; (6) Our method; (7) Ground truth.}
  \label{fig:comparision_different_pc}
\end{figure}

\begin{figure}[h!]
 \includegraphics[width=\linewidth]{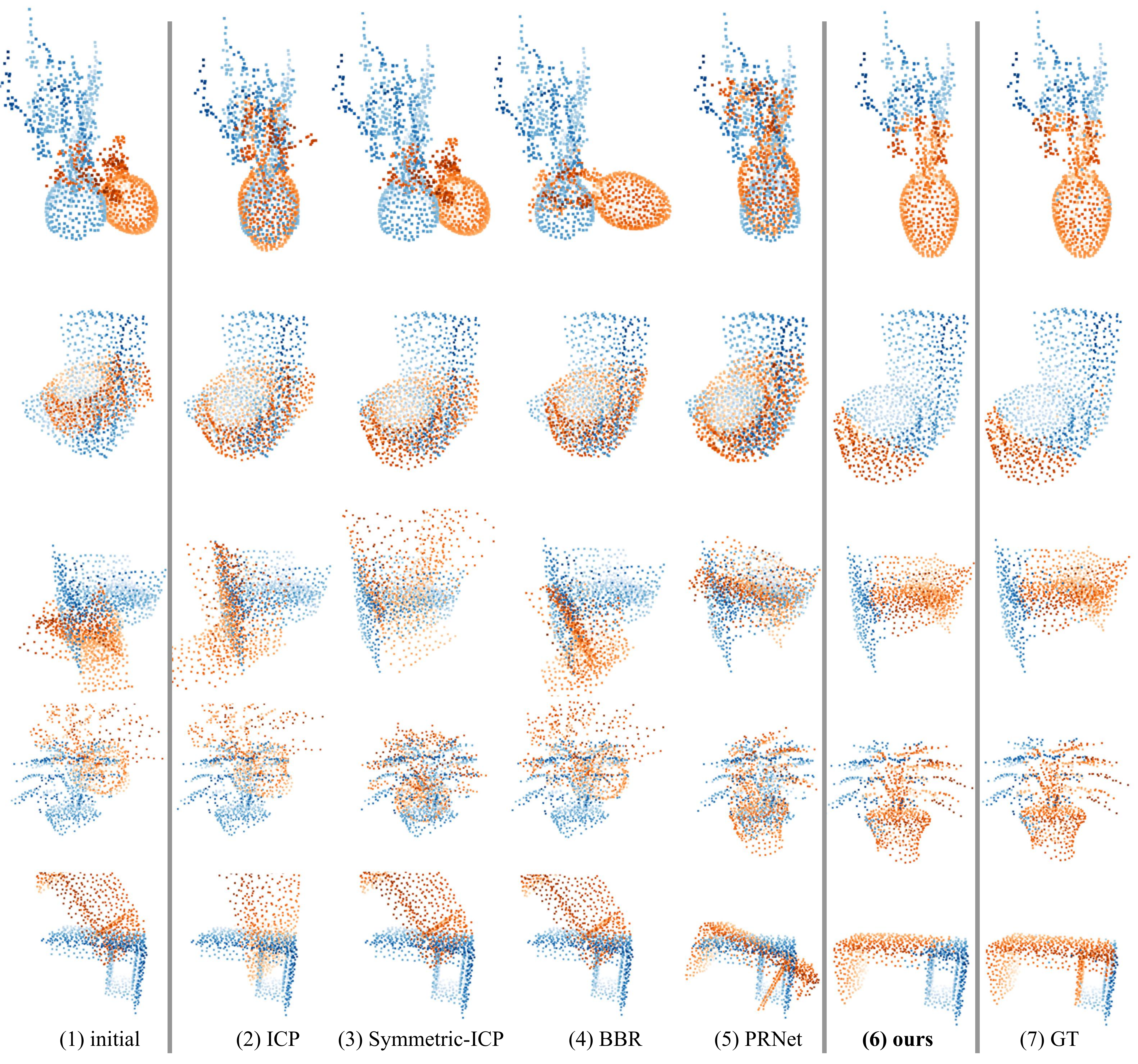}
\caption{\emph{Comparison with other methods - unseen point clouds test.} Each row contains different pairs of point clouds from the ModelNet40 dataset~\cite{ModelNet40}; (1) initial pose of the point clouds before registration; (2) ICP \cite{ICP}; (3) Symmetric-ICP~\cite{Rusinkiewicz:2019:ASO}; (4) BBR~\cite{drory2020best}; (5) PRNet~\cite{PRNet}; (6) Our method; (7) Ground truth.}
  \label{fig:comparision_unseen_pc}
\end{figure}

In Figure \ref{fig:comparision_unseen_pc} a visual comparison between our method and other methods is shown. The results are taken from the unseen point clouds experiment on the ModelNet40~\cite{ModelNet40} dataset (Section~\ref{modelnet40}). Figure \ref{fig:comparision_different_pc} shows results from the different sampling experiment.

\end{appendices}

\end{document}